\documentclass[preprint,12pt,authoryear]{elsarticle}

\usepackage{amssymb}
\usepackage{amsmath}
\usepackage[inline]{enumitem}
\usepackage{booktabs}
\usepackage{multirow}
\usepackage{graphicx}
\usepackage{subcaption}
\usepackage[margin=1in]{geometry}
\usepackage{caption}
\usepackage{makecell}
\usepackage{placeins}
\usepackage{tabularx}
\usepackage{bm}
\usepackage{xcolor}
\usepackage{hyperref}
\usepackage[capitalise]{cleveref}
\usepackage[acronym]{glossaries}
\usepackage{diagbox}
\usepackage{nth}

\crefname{appendix}{Appendix}{Appendices}
\Crefname{appendix}{Appendix}{Appendices}
\crefname{subappendix}{Appendix}{Appendices}
\Crefname{subappendix}{Appendix}{Appendices}
\crefname{subsubappendix}{Appendix}{Appendices}
\Crefname{subsubappendix}{Appendix}{Appendices}

\newacronym{ml}{ML}{Machine Learning}
\newacronym{dcm}{DCM}{Discrete Choice Models}
\newacronym{rum}{RUM}{Random Utility Models}
\newacronym{rumboost}{RUMBoost}{Random Utility Models with gradient Boosting}
\newacronym{gbdt}{GBDT}{Gradient Boosting Decision Trees}
\newacronym{paramboost}{ParamBoost}{Parametric Gradient Boosting}
\newacronym{fem}{FEM}{Functional Effects Models}
\newacronym{mnl}{MNL}{MultiNomial Logit}
\newacronym{mxl}{MXL}{MiXed Logit}
\newacronym{iclv}{ICLV}{Integrated Choice and Latent Variables}
\newacronym{nl}{NL}{Nested Logit}
\newacronym{cnl}{CNL}{Cross-Nested Logit}
\newacronym{ol}{OL}{Ordinal Logit}
\newacronym{lc}{LC}{Latent Class}
\newacronym{dl}{DL}{Dynamic Logit}
\newacronym{bp}{BP}{Binary Probit}
\newacronym{asc}{ASC}{Alternative-Specific Constant}
\newacronym{cel}{CEL}{Cross-Entropy Loss}
\newacronym{rf}{RF}{Random Forest}
\newacronym{mle}{MLE}{Maximum Likelihood Estimation}
\newacronym{vot}{VoT}{Value of Time}
\newacronym{wtp}{WTP}{Willingness-to-Pay}
\newacronym{nn}{NN}{Neural Network}
\newacronym{dnn}{DNN}{Deep Neural Network}
\newacronym{los}{LoS}{Level-of-Service}
\newacronym{ev}{EV}{Extreme Value}

\newcommand{\pb}[2]{\mbox{ParamB-$O^{#1}$-$C^{#2}$}}

\newif\ifshowtodos \showtodostrue

\newlist{inline}{enumerate*}{1}
\setlist[inline,1]{label=(\roman*)}

\journal{Transportation Research Part B: Methodological}

\begin{document}
\sloppy

\begin{frontmatter}

\title{A unified interpretable choice modelling framework with non-linear utility functions}

%% \author[ucl]{...}
%% \affiliation[ucl]{organization={...}, city={London}, country={United Kingdom}}

\author[inst1]{Nicolas Salvadé}
\ead{nicolas.salvade.22@ucl.ac.uk}
\affiliation[inst1]{organization={Department of Civil, Environmental and Geomatic Engineering, University College London},%Department and Organization
           addressline={Gower Street}, 
            city={London},
            postcode={WC1E 6BT}, 
            country={United Kingdom}}

\author[inst1]{Tim Hillel\corref{cor1}}
\cortext[cor1]{Corresponding author}
\ead{tim.hillel@ucl.ac.uk}

\begin{abstract}
Hybrid models that combine machine learning with random utility theory can learn non-linear utility functions from data while maintaining interpretability and behavioural consistency. However, hybrid modelling approaches in the literature predominantly rely on simple model structures (e.g. \acrshort{mnl}) and, to the best of our knowledge, no general, interpretable hybrid modelling frameworks exist that can be extended to arbitrary model structures. Building on previous work investigating non-linear utility functions with gradient boosting, we present here a general choice modelling framework and show that it extends to a broad class of discrete choice models. We describe a choice model by four components:
\begin{inline}
    \item the coefficient function that defines how each feature enters the utility;
    \item the aggregation function, combining the coefficient functions;
    \item the probability function obtained from an assumption on the error term, mapping utilities to choice probabilities; and
    \item the loss function, which quantifies the error between predicted and observed choice probabilities. 
\end{inline}
We then use this framework to compare linear-in-parameter models with hybrid \acrshort{ml}--\acrshort{dcm}s and, in doing so, show how gradient boosting can model fully interpretable non-linear coefficient functions for any model with differentiable probability and loss functions. Building on recent work on parametric \acrshort{ml} models, we show how the framework lets the modeller constrain the degree of smoothness, monotonicity, and convexity of the utility as required for different applications. Within this framework, we implement nine models: multinomial, mixed, nested, cross-nested, ordinal, latent class and dynamic logit models, binary probit model, and an integrated choice and latent variable model, testing four different functional forms of the utility for each model. We evaluate these models against conventional linear-in-parameters choice models on two real-world mode-choice datasets. The results show that gradient-boosted utility models achieve a lower test cross-entropy than linear-in-parameters baselines estimated with Biogeme in 35 of 36 comparisons. Finally, we investigate how the degree of smoothness of the coefficient functions affects both predictive performance and interpretability, finding that smoother utilities consistently overfit less than those with discontinuities.
\end{abstract}

\begin{highlights}
\item Introduces a unified choice modelling framework based on gradient-boosted utilities
\item The framework extends to any choice model with a differentiable probability and loss function
\item We implement nine choice models within the framework
\item Boosted models beat linear-in-parameters baselines in 35 of 36 comparisons
\item Maintains interpretability with different degrees of smoothness in utility functions
\end{highlights}

\begin{keyword}
Machine Learning \sep Gradient Boosting \sep Discrete Choice Models \sep Random Utility \sep Piecewise Polynomial Utility
\end{keyword}

\end{frontmatter}

%% ===================================================================
\section{Introduction}\label{sec:intro}

\acrfull{dcm} based on the random utility theory have been the standard tool for modelling individual choice behaviour for more than fifty years. Their advantages lie in interpretable parameters, behavioural consistency and robust estimation. However, these properties depend largely on a linear-in-parameters utility function that the modeller must specify. Over the past decade, a growing body of literature has combined flexible \acrfull{ml} methods with \acrshort{dcm}s to relax this linearity assumption.%, typically improving predictive performance at some cost of interpretability.

The flexibility of hybrid \acrshort{ml}--\acrshort{dcm} approaches has been consistently shown to improve out-of-sample predictive performance relative to conventional linear-in-parameter choice models (see, e.g. \cite{WANG2020102701, sifringer2020enhancing, han2022neural, wong2019reslogit, salvade2025rumboost, salvade2026paramboost}). In addition, they retain some degree of behavioural interpretability (typically via constraints on variable interactions in the model) compared to pure black-box \acrshort{ml} classification models. Despite these advantages, we identify two general limitations across the body of existing work that hinder their widespread application for choice modelling:
\begin{enumerate}
    \item non-linearity is often achieved at the cost of a closed-form utility specification, such that the utility function is not observed over the full input space; and 
    \item hybrid \acrshort{ml}-\acrshort{dcm} research has typically focused on simple model specifications based on the \acrfull{mnl}, limiting the range of choice structures and behavioural constructs that can be implemented. 
\end{enumerate}

The former limitation - the absence of a closed-form utility function - means that consistent marginal effects cannot be observed directly from the model. This is problematic as these effects are used to compute key behavioural indicators such as the \acrfull{wtp} and the \acrfull{vot}. Indeed, without a closed-form expression, only point elasticities can be approximated at data points, and their variations across the input space offer no guarantee of validity. The latter limitation - the over-reliance on an \acrshort{mnl}-like model structure - similarly limits the capabilities of hybrid choice models. More complex choice constructs are necessary to accurately model complex behavioural phenomena, such as dynamic or ordinal choices, nested or cross-nested structures, distributed parameters or latent variables, and other assumptions on the error term. However, researchers have investigated these complex choice constructs only sporadically within hybrid \acrshort{ml}-\acrshort{dcm}s, highlighting the lack of a unified framework for representing complex behaviours in these models.

Among these hybrid approaches, the \acrshort{rumboost} model family \citep{salvade2025rumboost, salvade2025functional, salvade2026paramboost} keeps the utility function fully interpretable while improving predictive performance. Using an extension of the \acrfull{gbdt} algorithm, the models output non-linear utility functions with a fully known functional form. They can also learn distributed parameters suitable for inference. Although the models have been applied to complex choice models such as the nested logit model or static models with distributed parameters in panel data, they have not been tested systematically on a wide variety of complex choice constructs.

This paper presents a novel choice modelling framework that covers both non-linear and linear-in-parameters utility functions that we use to derive a wide variety of choice models. Our framework defines a choice model by four components:
\begin{inline}
    \item the coefficient function, mapping how a feature enters the utility function;
    \item the aggregation function, combining the coefficients into the utility function;
    \item the probability function, mapping the utilities to choice probabilities following an assumption on the distribution of the error term; and
    \item the loss function, quantifying how much choice predictions differ from the ones observed in the data.
\end{inline}

\acrshort{ml} can learn the coefficient functions from the data, and it interacts with the other choice model components only through the derivatives of the loss function with respect to the coefficient function. Any model with a differentiable aggregation, probability and loss function can therefore be estimated with \acrshort{ml}. In practice, we rely on \acrshort{gbdt} because prior work demonstrated glass-box interpretability with closed-form functional expressions \citep{salvade2025rumboost, salvade2025functional, salvade2026paramboost}. This motivates our choice of \acrshort{gbdt} in the framework. In theory, any gradient-based \acrshort{ml} model can learn the coefficient functions. For instance, we could learn coefficient functions with \acrfull{nn}s, but they would not offer closed-form functional forms of the utility function.

The main contributions of this paper are:
\begin{enumerate}
    \item a unified choice modelling framework to automatically learn from the data non-linear utility functions, distributed parameters suitable for inference and variable interactions;
    \item implementations of nine \acrshort{dcm}s within this framework: the \acrshort{mnl}, \acrfull{mxl}, \acrfull{iclv}, \acrfull{nl}, \acrfull{cnl}, \acrfull{ol}, \acrfull{lc}, \acrfull{dl} and \acrfull{bp} models. Among these models, the \acrshort{iclv}, \acrshort{cnl}, \acrshort{lc}, \acrshort{dl} and \acrshort{bp} implementations are novel to this work;
    \item a benchmark of four model variants obtained with the framework against a linear-in-parameter utility baseline estimated with Biogeme \citep{bierlaire2023short}; and
    \item an open source implementation of the framework using the \verb|JAX| \citep{jax2018github} Python library\footnote{https://github.com/big-ucl/a-nonlinear-choice-modelling-framework and https://github.com/big-ucl/paramboost}.
\end{enumerate}

\Cref{sec:background} reviews related work and the RUMBoost model family. \Cref{sec:framework} then introduces the framework, and \cref{sec:models} shows how the nine models fit within it. \Cref{sec:experiments} reports the empirical evaluation, and \cref{sec:conclusion} discusses the results, their limitations and future work.
%% ===================================================================
\section{Background}\label{sec:background}

\subsection{Specification of utility functions in discrete choice models}\label{sec:bg_spec}

While non-linearities in \acrshort{dcm}s are possible, the modeller must introduce them. This time-consuming, cumbersome task cannot guarantee optimal transformations because of the size of the search space. Several studies have attempted to alleviate this issue by providing assisted specification of a choice model to be estimated through maximum likelihood estimation. \cite{hillel2019weak} use the output of a \acrshort{gbdt} ensemble and \cite{shiftan2020utilizing} use a \acrfull{rf} ensemble to specify box-cox transforms and variable interactions for an \acrshort{mnl}. \cite{salvade2025assisted} use the output of a \acrshort{gbdt} algorithm with linear trees to inform the thresholds of a piecewise linear \acrshort{mnl}. In contrast, \cite{ke2023discrete} incorporate the potential thresholds directly into the maximum likelihood estimation. \cite{paz2019specification} use an optimisation framework to assist the specification of an \acrshort{mxl} model. Furthermore, \cite{hernandez2023data} use association rules and an \acrshort{rf} algorithm to help specify a portfolio model, showing improved model fit. Finally, researchers also use more general approaches to specify a wider variety of choice models, including a variant of the neighbourhood search algorithm \citep{ortelli2021assisted}, a grammar-based approach \citep{yahiagrammar}, and reinforcement learning \citep{nova2025improving, nova2026multitask}. All of these studies attempt to identify an improved model specification for conventional choice models by exploring the search space through optimisation or machine learning. However, they rely on either pre-specified model specifications or the modeller's observation of non-linearities and variable interactions.

\subsection{Hybrid \acrshort{ml}--\acrshort{dcm}s}\label{sec:bg_hybrid}

In contrast to assisted specification methods, researchers have applied restricted \acrshort{ml} models to alleviate the rigid structure of discrete choice models and improve their flexibility and predictive performance. Most research focuses on deep learning. For instance, \cite{WANG2020102701} use a \acrfull{dnn} to learn alternative-specific utilities, and \cite{sifringer2020enhancing} use the L-MNL model to separate the utility function into an interpretable linear-in-parameter part for trip attributes, while feeding socio-demographic characteristics into a fully connected, black-box \acrshort{nn}. The outputs of the two parts are then summed to obtain the deterministic utilities. \cite{han2022neural} extend this idea in TasteNet-MNL to learn distributed parameters from socio-demographic characteristics using \acrshort{nn}, while \cite{aouad2022representing} and \cite{bagheri2025rum} define a general framework for representing random utility models with \acrshort{nn}s. \cite{hernandez2024economically} derive an economically consistent \acrshort{dcm} using \acrshort{nn}, and \cite{yun2026neural} apply a neural bayesian learning structure to learn \acrshort{dcm}s. \cite{das2026utility} constrain the gradients of a \acrshort{nn} so that the utility gradients are monotonic, and \cite{wong2019reslogit} learn cross-alternative effects using a ResNet model, deriving the ResLogit model. 

Researchers have also experimented with other \acrshort{ml} algorithms. For instance, \cite{villarraga2025designing} use a graph convolutional network to incorporate network structure within a \acrshort{dcm} and \cite{VILLARRAGA2025103132} adapt Gaussian processes for \acrshort{dcm}. \cite{arkoudi2023combining, baek5765940generalized, wang2026embedding} all incorporate embeddings within \acrshort{dcm}s. Using a Lattice Network, \cite{kim2024new} learn partially monotonic non-linear utility functions, while \cite{martin2021revisiting} use generalised kernel logistic regression to learn hybrid choice models. \cite{krueger2022stated} use the Choquet integral to investigate pairwise variable interaction and \cite{aboutaleb2022theory} learn non-linear utility functions using sum-of-squares polynomials. Finally, researchers also adapted \acrshort{gbdt} for \acrshort{rum} with the \acrshort{rumboost} model family \citep{salvade2025functional, salvade2025rumboost, salvade2026paramboost}.
The main difference among these methods is the level of constraints applied to the algorithm, i.e., feature interactions, monotonicity or shape constraints, and model structure (e.g., the possibility of alternative-specific features).

\subsection{The RUMBoost model family}\label{sec:bg_rumboost}

The \acrshort{rumboost} model family can automatically learn a non-linear utility function and distributed parameters while providing a functional form for the utility function. The family comprises three models: 
\begin{inline}
    \item \acrshort{rumboost};
    \item \acrfull{fem}; and
    \item \acrfull{paramboost}.
\end{inline}
The latter two models extend the first, with \acrshort{fem} learning distributed parameters by mapping socio-demographic characteristics to individual-specific trip parameters using an ensemble of regression trees, and \acrshort{paramboost} outputting a non-linear utility function with a fully known functional form. Both models are obtained by boosting at the parameter level. The difference is that in \acrshort{fem} the authors split on socio-demographic characteristics to obtain individual-specific parameters, which will be multiplied by a trip attribute, while in \acrshort{paramboost}, the authors split on the variable itself to obtain directly a, at most, cubic piece-wise utility function with degree of smoothness \textit{a priori} specified by the modeller.

In this paper, we present a unified choice modelling framework that automatically learns non-linear utility and distributed parameters from the data. \Cref{sec:framework} presents the framework, and \cref{sec:models} shows how the framework can generalise to a wide variety of choice models.

%% ===================================================================
\section{A framework for non-linear choice models}\label{sec:framework}

\begin{figure}[htbp]
    \centering
    \includegraphics[width=\textwidth]{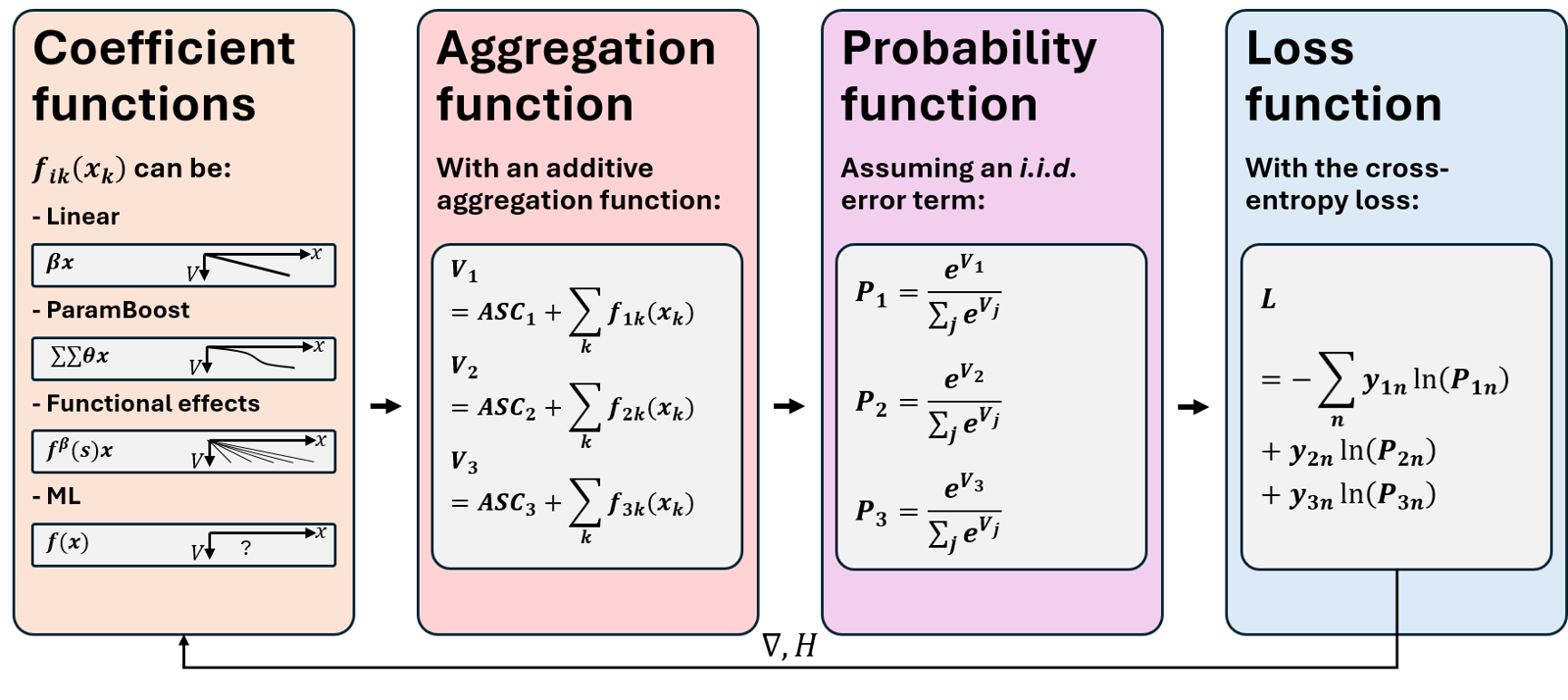}

    \caption{Diagram of the framework, assuming three alternatives with the \acrshort{mnl} model structure as example.}
    \label{fig:diagram}
\end{figure}

In this section, we define a choice model through four components (\cref{sec:components}), describe how gradient boosting learns coefficient functions (\cref{sec:learning_f}), and show how to apply this approach to any differentiable choice models (\cref{sec:generalisation_framework}). The \acrshort{mnl} serves as a running example; every model in \cref{sec:models} is obtained from it by changing one or more components. \Cref{fig:diagram} summarises the framework.

\subsection{Four components of a choice model}\label{sec:components}

\paragraph{Coefficient function} The coefficient function $f$ defines how each feature enters the utility. In the \acrshort{mnl} the function is linear:
\begin{equation} \label{eq:coefficient_function}
    f_{ik}(x_k) = \beta_{ik} x_k
\end{equation}
where $\beta_{ik}$ is the parameter associated with feature~$k$ in the utility of alternative~$i$, typically learnt through \acrfull{mle}.
\Cref{sec:learning_f} replaces the linear coefficients with non-linear functions learnt using \acrshort{ml} methodologies.

\paragraph{Aggregation function} The aggregation operator $\mathcal{H}$ combines the contributions of individual features into the deterministic utility of each alternative~$i$:
\begin{equation} \label{eq:aggregation}
    V_i = \mathcal{H}\big(f_{ik}(x_k)\big).
\end{equation}
In random utility models, $\mathcal{H}$ is almost always a sum, although non-additive forms such as Cobb--Douglas utility functions exist \citep{zellner1966specification}. The framework applies to any aggregation operator. However, all examples in this paper use additive aggregation, as is standard in the discrete choice literature. With additive aggregation and a linear coefficient function, this gives:
\begin{equation} \label{eq:linear_utility}
    V_i = \text{ASC}_i + \sum_{k=1}^{K}\beta_{ik}x_{k},
\end{equation}
where $\text{ASC}_i$ is the alternative-specific constant of alternative~$i$ and $K$ the number of features.

\paragraph{Probability function} The random utility is composed of the deterministic utility and an error term $\varepsilon$ capturing measurement error and unobserved influences on choice:
\begin{equation} \label{eq:random_utility}
    U_i = V_i + \varepsilon_i.
\end{equation}
An individual chooses alternative~$i$ if $U_i \geq U_j$ for all $j$, such that
\begin{equation} \label{eq:probability}
    P_i = \Pr\left(\varepsilon_j - \varepsilon_i \leq V_i - V_j \;\; \forall j \neq i\right).
\end{equation}
The assumed distribution of the differences of error terms defines the probability function. In the \acrshort{mnl}, we assume that the error term is extreme value type I distributed, which gives the logistic, or softmax function:
\begin{equation} \label{eq:mnl_probability}
    P_i = \frac{e^{V_i}}{\sum_{j=1}^{J}e^{V_j}},
\end{equation}
where $J$ is the number of alternatives.

\paragraph{Loss function} The parameters are chosen to minimise a loss $\ell(P, y)$ that compares the predicted probabilities with the observed choices $y$. While the framework can accommodate any differentiable loss function, for discrete choice models including the MNL we use the  \acrfull{cel}:
\begin{equation} \label{eq:cel}
    \mathcal{L} = \frac{1}{N}\sum_{n=1}^{N} \ell(P_n,y_n)= -\frac{1}{N}\sum_{n=1}^{N}\sum_{j=1}^{J} y_{jn}\ln P_{jn},
\end{equation}
where $N$ is the number of observations and $y_{jn} = 1$ if individual $n$ chose alternative $j$ and $0$ otherwise. Minimising \cref{eq:cel} is equivalent to \acrshort{mle}. \acrshort{cel} is the default metric reported for each model in \cref{sec:experiments}.

\subsection{Learning coefficient functions with \acrshort{ml}}\label{sec:learning_f}

Our generalised non-linear utility framework replaces the linear coefficients in \cref{eq:coefficient_function} with non-linear functions learnt automatically through \acrshort{ml}. Whilst the framework allows for any non-linear \acrshort{ml} model estimated through gradient descent, including \acrshort{nn}s, we focus here on models with \emph{intrinsic interpretability}, where the full functional form is known over the possible range of input values. Specifically, we use gradient boosting to learn coefficients as
\begin{inline}
    \item piece-wise polynomial functions (such that the coefficient is a function of the corresponding feature~$x_k$), adopting the prior work of \cite{salvade2026paramboost}; or
    \item individual-specific constants dependent on a respondent's socio-demographic characteristics~$\bm{s}_n$, adopting the prior work of \cite{salvade2025functional}.
\end{inline}

\subsubsection{Non-linear parametric coefficient functions} \label{subsubsec:gb_coeff_function}

Classical \emph{black-box} multinomial classification uses gradient boosting to learn an ensemble of decision trees~$F_i$ for each class~$i$ as a function of all available features~$\bm{x}$. The ensemble outputs~$z_i$ are then used to compute class probabilities~$P_i$ via the softmax function. The split points and leaf values are learnt directly from the first- and second-order derivatives of the (cross-entropy) loss with respect to these outputs \citep{friedman2001greedy, chen2016xgboost, ke2017lightgbm}. 

Instead, we adopt the methodology of \acrshort{paramboost} \citep{salvade2026paramboost} which learns an ensemble of regression trees for each \emph{coefficient}, as a function of \emph{only} the corresponding feature(s)~$x_k$. By including polynomial transformations of $x_k$ in the ensemble, this supports fully glass-box non-linear coefficient functions. We provide a brief overview of the core methodology, including monotonicity and shape constraints, below, and direct the reader to the full paper for further detail.

\acrshort{paramboost} \citep{salvade2026paramboost} presents a parametric \acrshort{ml} formulation where the coefficient function of each feature is learnt as a piecewise polynomial of maximum degree $D$ over $B$ intervals using a gradient-boosted ensemble:
\begin{equation} \label{eq:piecewise_polynomial}
    f_{ik}(x_k) = \sum_{b=1}^{B}\sum_{d=0}^{D}\theta_{ikbd}\, {x_{kb}}^{d},
\end{equation}
with:
\begin{equation} \label{eq:piecewise_basis}
    x_{k1} = \min(x_k, u_1), \qquad
    x_{kb} =
    \begin{cases}
    0 & \text{if } x_{k} < u_{b-1}, \\
    x_{k} - u_{b-1} & \text{if } u_{b-1} \leq x_{k} < u_{b}, \\
    u_b - u_{b-1} & \text{if } x_{k} \geq u_{b},
    \end{cases}
    \quad b = 2, \dots, B,
\end{equation}
where $u_1 < \dots < u_{B-1}$ are the interval bounds ($u_B = \infty$) and $\theta_{ikbd}$ is the parameter associated with alternative~$i$, feature~$k$, interval~$b$ and degree~$d$. In other words, each linear parameter $\beta_{ik}$ in a linear-in-parameters \acrshort{dcm} is replaced by up to $ B \times D$ parameters, representing a piecewise polynomial function. In \acrshort{paramboost}, the bounds ($u_b$) are determined automatically before model fitting, typically including the total number of intervals $ B$, but the modeller can also constrain them.

\acrshort{paramboost} uses histogram-based \acrshort{gbdt}: it discretises each feature into bins, whose edges serve as feasible values for the interval bounds $u_b$. At each boosting iteration, every bin edge of every feature and degree is a candidate split point. The split that most reduces the loss is selected, and all parameters~$\theta_{ikbd}$ on either side are updated with the corresponding leaf value, a constant representing the coefficient of the degree~$d$ polynomial. This means that, at each iteration, only two leaf values need to be computed to update all parameters, unlike other gradient-based optimisation techniques, which require computing new values for each parameter.

A split on the degree-$d$ terms creates a difference in those terms between neighbouring intervals. Smoothness up to any required degree $S$ (i.e. $C^S$) can be enforced by disallowing splits on monomials of degree $d\leq S$ and instead learning a single coefficient~$\beta_{ikd}$ over the full input range for the corresponding feature. For example, a continuous ($C^0$) utility requires that the constant terms ($d = 0$) are never split, since a split would introduce a discontinuous step at the split point. Combining the maximum degree $D$ with the smoothness level $S$ gives the family in \cref{tab:piecewise_poly}, which contains \acrshort{rumboost} ($D = 0$, discontinuous) and the linear-in-parameters \acrshort{mnl} ($D = 1$ with a single parameter) as special cases. We denote a variant by \pb{D}{S}, with $C^{-1}$ indicating a discontinuous function, and use the four variants in bold throughout the paper for conciseness. We choose these variants because they represent the full range of utility smoothness that \acrshort{paramboost} can provide. Note that power or logarithmic transformations, commonly implemented in choice models through the Box--Cox transform, cannot be explicitly represented with \acrshort{paramboost}, but similar functional forms can be approximated through piecewise cubic functions.

\begin{table}[htbp]
\centering
\caption{Piecewise polynomial utility functions by maximum degree and continuity. The four variants used in this paper are in bold.}
\label{tab:piecewise_poly}
\footnotesize
\renewcommand{\arraystretch}{1.3}
\begin{tabular}{lllll}
\toprule
\diagbox{Cont.}{Max deg.} & 0 & 1 & 2 & 3 \\
\midrule
$C^{-1}$ & \makecell[l]{\textbf{RUMBoost}\\ (\pb{0}{-1})} & Disc.\ linear & Disc.\ quadratic & \makecell[l]{\textbf{Disc.\ cubic}\\ (\pb{3}{-1})} \\
$C^{0}$  & Constant & \makecell[l]{\textbf{Linear splines}\\ (\pb{1}{0})} & $C^0$ quadratic & $C^0$ cubic \\
$C^{1}$  & -- & MNL/MXL & Quadratic splines & $C^1$ cubic \\
$C^{2}$  & -- & -- & Quadratic & \makecell[l]{\textbf{Cubic splines}\\ (\pb{3}{2})} \\
\bottomrule
\end{tabular}
\end{table}

We can impose monotonicity and convexity constraints on each feature \citep{salvade2026paramboost}. In this paper, trip \acrfull{los} attributes, such as travel time and cost, are constrained to negative monotonicity, so an increase in the attribute always decreases utility. Where applicable (i.e., for the ParamB-$O^1$-$C^0$, ParamB-$O^3$-$C^-1$, and ParamB-$O^3-C^2$ models), we also constrain the coefficient functions for \acrshort{los} attributes to be convex, so they have diminishing marginal disutility (\cref{sec:setup}).

\subsubsection{Inter-individual heterogeneity} \label{subsubsec:heterogeneity}

The non-linear coefficient functions in \cref{subsubsec:gb_coeff_function} are shared across the whole population. To represent systematic heterogeneity as an alternative to mixture models, we instead generate individual-specific linear parameters $\beta_n$ as a function of the socio-demographic characteristics $\mathbf{s}_n$, following the approach introduced by \cite{han2022neural}, and extended to gradient-boosted utility models in \cite{salvade2025functional}.
The coefficient~$\beta$ for alternative~$i$, feature~$k$ and individual~$n$ is given as a function of the socio-demographic characteristics $\mathbf{s}_n$,
\begin{equation} \label{eq:functional_effect}
    \beta_{ikn} = f_{ik}^{\beta}(\mathbf{s}_n)
\end{equation}
where $f_{ik}^{\beta}$ is an ensemble of regression trees. Monotonicity can be constrained through a ReLU transformation on the output, which ensures all $\beta_{kn}$ are strictly positive or negative as required:
\begin{equation}
    \beta_{kn}=\pm\text{ReLU}(f^\beta_k(\mathbf{s}_n)) 
\end{equation}
Note that an ensemble can also output a utility term directly, $f^{V}(\mathbf{s})$, as in \acrshort{rumboost}, if the derivatives are computed in the utility space; \cref{sec:heterogeneity} uses this to learn a latent variable from the socio-demographic characteristics.

\subsection{A unified choice modelling framework}\label{sec:generalisation_framework}

The coefficient functions in \cref{sec:learning_f} are learnt to minimise a loss function. We adopt the approach introduced in \cite{friedman2001greedy} (see also \cite{chen2016xgboost, ke2017lightgbm} for modern implementations of the algorithm), which optimises both splits and leaf values from the first- and second-order derivatives of the loss~$\ell$ with respect to each ensemble output $z$. Through application of the chain rule, this gives, for the first-order derivatives:

\begin{equation}\label{eq:first-order-chain-rule}
    \frac{\partial \ell}{\partial z} = \frac{\partial \ell}{\partial P_i} \frac{\partial P_i}{\partial V_j} \frac{\partial V_j}{\partial{z}}, \quad \forall j=1, \dots, J.
\end{equation}

And for the second-order derivatives, using the chain rule with the product rule:
\begin{equation}\label{eq:second-order-chain-rule}
    \frac{\partial^2 \ell}{\partial z^2} = \frac{\partial^2 \ell}{\partial P_i^2} \left(\frac{\partial P_i}{\partial V_j}\right)^2 \left(\frac{\partial V_j}{\partial{z}}\right)^2 + \frac{\partial \ell}{\partial P_i} \frac{\partial^2 P_i}{\partial V_j^2} \left(\frac{\partial V_j}{\partial{z}}\right)^2 +\frac{\partial \ell}{\partial P_i} \frac{\partial P_i}{\partial V_j} \frac{\partial^2 V_j}{\partial{z}^2},  \quad \forall j=1, \dots, J.
\end{equation}

Each term in \cref{eq:first-order-chain-rule} and \cref{eq:second-order-chain-rule} relates to one of the remaining components of our choice modelling framework: $\frac{\partial V_j}{\partial z}$ and $\frac{\partial^2 V_j}{\partial z^2}$ are determined by the utility aggregation function; $\frac{\partial P_i}{\partial V_j}$ and $\frac{\partial^2 P_i}{\partial V_j^2}$ are determined by the probability function; and $\frac{\partial \ell}{\partial P_i}$ and $\frac{\partial^2 \ell}{\partial P_i^2}$ are determined by the loss function\footnote{Note that in the \acrshort{gbdt} algorithm, the second derivatives are a diagonal approximation of the Hessian, which can be interpreted as assuming independence between alternatives.}.

\paragraph{Utility aggregation function} 
The restriction to additive aggregation function makes calculation of $\frac{\partial V_j}{\partial z}$ and $\frac{\partial^2 V_j}{\partial z^2}$ straightforward, depending on how what $z$ represents. \Cref{tab:aggregation_fct_derivatives} summarises these derivatives by coefficient function type.
\begin{table}[htbp]
    \centering
    \begin{tabular}{lcc}
    \toprule
        $z$ & $\frac{\partial V_j}{\partial z}$ & $\frac{\partial^2 V_j}{\partial z^2}$  \\ \midrule
        $\theta_{jkbd}$ & $(x_{kb})^d$ & $(x_{kb})^{2d}$  \\
        $\beta_{jkn}$ & $x_k$ & $x_k^2$ \\
        \bottomrule
    \end{tabular}
    \caption{Aggregation function derivatives, with respect to coefficient function type.}
    \label{tab:aggregation_fct_derivatives}
\end{table}

Note that in the case that $d=0$, both $\frac{\partial V_j}{\partial z}=1$ and $\frac{\partial^2 V_j}{\partial z^2}=1$, simplifying to a utility (or functional) space \acrshort{gbdt} algorithm.

\paragraph{Probability function}

The use of a type 1 \acrfull{ev} error term results in the mathematically convenient \acrshort{mnl} probability function, where both $\frac{\partial P_i}{\partial V_j}$ and $\frac{\partial^2 P_i}{\partial V_j^2}$ have a simple form ($P_i (1-P_i)$ and $P_i(1-P_i)(1-2P_i)$ respectively if $i = j$, $P_jP_i$ and $P_jP_i$ respectively if $i \neq j$).
When an assumption on the error term does not yield closed-form expressions for the derivatives, automatic differentiation, as implemented in deep learning libraries such as \verb|PyTorch| \citep{10.1145/3620665.3640366} and \verb|Jax| \citep{jax2018github}, can be used instead.

\paragraph{Loss function}
Again, the use of the \acrshort{cel} function, akin \acrshort{mle}, leads to mathematically convenient derivatives, where $\frac{\partial \ell}{\partial P_i} = \frac{1}{P_i}$ and $\frac{\partial^2 \ell}{\partial P_i^2} = -\frac{1}{P_i^2}$. For other loss functions, we compute these derivatives analytically or using automatic differentiation.

\paragraph{Scalar parameters}
The above analysis covers all parameters that are functions of input features. However, many choice models also rely on scalar parameters that are not functions of the features, e.g.: the nest parameters $\mu$ and membership parameters $\alpha$ of nested and cross-nested logit models, the class shares $\pi$ of latent class models, the thresholds $\tau$ of ordinal models, and the measurement-equation parameters of \acrshort{iclv} models. We achieve this by treating all observations as belonging to a single leaf node and calculating the leaf value using the first- and second-order gradients from \cref{eq:first-order-chain-rule} and \cref{eq:second-order-chain-rule}. This is equivalent to a quasi-Newton method with a diagonal Hessian approximation. Note that these scalar parameters are updated sequentially at each boosting iteration; we first find the optimal split point and leaf values, add them to the coefficient functions accordingly, and then update the scalar parameters.

\paragraph{Variable interactions}
While this paper does not directly demonstrate this, the framework can be used to automatically learn variable interactions while remaining interpretable up to two variable interactions. This is only limited to \pb{0}{-1}, but monotonicity interactions can still be incorporated.

%% ===================================================================
\section{Choice models within the framework}\label{sec:models}

This section introduces eight additional choice model specifications within the framework, each obtained from the \acrshort{mnl} of \cref{sec:components} by changing one or more components. \Cref{tab:model_summary} summarises the changes. The models are ordered by how far they differ from the \acrshort{mnl}:
\begin{inline}
    \item a change to the utility structure only (\cref{sec:dl});
    \item to the probability function only (\cref{sec:prob});
    \item to both (\cref{sec:ol_lc}); and
    \item two models whose linear-in-parameters versions rely on simulation and whose \acrshort{paramboost} counterparts therefore differ from the baseline (\cref{sec:heterogeneity}).
\end{inline}
Unless stated otherwise, \acrshort{paramboost} models use the coefficient functions of \cref{sec:learning_f} for every trip attribute, and the baselines use linear coefficients. These eight model formulations are not exhaustive; they are intended to show how the framework can be adapted and extended to varied model specifications. The framework and the implemented models are open source and freely available on GitHub\footnote{https://github.com/big-ucl/a-nonlinear-choice-modelling-framework and https://github.com/big-ucl/paramboost}.

\begin{table}[htbp]
\centering
\caption{How each choice model departs from the \acrshort{mnl}, and how its \acrshort{paramboost} counterpart differs from the linear-in-parameters baseline.}
\label{tab:model_summary}
\tiny
\begin{tabularx}{\textwidth}{l l >{\raggedright\arraybackslash}X >{\raggedright\arraybackslash}X l c}
\toprule
Model & Full name & Component(s) changed relative to the MNL & ParamBoost-specific treatment & Data & New \\
\midrule
\acrshort{dl} & \acrlong{dl} & Utility: lagged-choice term & -- & Swissmetro & \checkmark \\
\acrshort{bp} & \acrlong{bp} & Probability: normal CDF of the utility difference & -- & Swissmetro, binary & \checkmark \\
\acrshort{nl} & \acrlong{nl} & Probability: nested GEV ($\mu$) & -- & Swissmetro & \\
\acrshort{cnl} & \acrlong{cnl} & Probability: cross-nested GEV ($\mu$, $\alpha$) & -- & Swissmetro & \checkmark \\
\acrshort{ol} & \acrlong{ol} & Utility: single latent utility; probability: ordered logit ($\tau$) & -- & Swissmetro, ordered & \\
\acrshort{lc} & \acrlong{lc} & Utility: class-specific; probability: mixture over classes ($\pi$) & -- & Swissmetro & \checkmark \\
\acrshort{mxl} & \acrlong{mxl} & Coefficients: random; loss: simulated & Coefficients learnt as functional effects; no simulation & Swissmetro & \\
\acrshort{iclv} & \acrlong{iclv} & Utility: latent variable; loss: joint with measurement model, simulated & Latent variable learnt as a functional effect; no simulation & Optima & \checkmark \\
\bottomrule
\end{tabularx}
\end{table}

\subsection{Changing the utility structure: dynamic logit}\label{sec:dl}

\paragraph{Dynamic logit}
Dynamic models describe sequential choices by capturing changes in the decision process over time \citep{bierlaire2024dynamic}. We formulate here a first-order Markov model, where the choice at time $t$ depends additionally on the previous choice at time $t-1$, representing myopic behaviour without agent effects:
\begin{equation} \label{eq:dl_utility}
    V_{int} = \text{ASC}_i + \sum_{k=1}^{K} f_{ik}(x_{knt}) + \eta_i\, y_{in,t-1},
\end{equation}
where $y_{in,t-1} = 1$ if individual $n$ chose alternative $i$ in choice situation $t-1$ and $0$ otherwise. We assume the previous choice captures the correlation between repeated choices of the same individual, so the error term is independent over $t$ and the probability function and loss are those of the \acrshort{mnl}, \cref{eq:mnl_probability} and \cref{eq:cel}. Because $y$ is binary, any learnt function of it reduces to a constant shift, so the boosted and linear versions differ only in the coefficients of the trip attributes. In \acrshort{paramboost}, $\eta$ is computed the same way as for binary variables, using an ensemble of regression trees of degree $d=0$ (since any degree $d>0$ provides the same information for a binary variable).

\subsection{Changing the probability function: binary probit, nested, cross-nested and ordinal logit}\label{sec:prob}

\paragraph{Binary probit} A binary probit model models binary outcomes and assumes the error term is normally distributed. Assuming normally distributed error terms changes only the probability function. For two alternatives, let $V = V_1 - V_2$ be the utility difference and normalise the variance of the error difference to one. Then
\begin{equation} \label{eq:bp_probability}
    P_1 = \Phi(V),
\end{equation}
where $\Phi$ is the standard normal cumulative distribution function. With the binary form of \cref{eq:cel}, the derivative needed in \cref{eq:first-order-chain-rule} is:
\begin{equation} \label{eq:bp_gradient}
    \frac{\partial P_n}{\partial V_n} = \frac{\phi(V_n)}{\Phi(V_n)}
\end{equation}
and in \cref{eq:second-order-chain-rule} is:
\begin{equation} \label{eq:bp_hess}
    \frac{\partial^2 P_n}{\partial V_n^2} = \frac{V_n\phi(V_n)\Phi(V_n) -\phi(V_n)^2}{\Phi(V_n)^2}
\end{equation}
where $\phi$ is the standard normal density function.

\paragraph{Nested logit} Nested logit models relax the independence of the \acrshort{mnl} error terms by grouping correlated alternatives into nests. With $M$ nests whose nest parameters are $\mu_m \geq 1$, the probability of choosing alternative $i$ in nest $m$ is:
\begin{equation} \label{eq:nl_probability}
    P_i = \frac{e^{\mu_m V_{i}}}{\sum_{j\in m}e^{\mu_m V_{j}}}\,\frac{e^{\tilde{V}_{m}}}{\sum_{p=1}^{M}e^{\tilde{V}_{p}}},
    \qquad
    \tilde{V}_{m} = \frac{1}{\mu_m} \ln \sum_{j \in m} e^{\mu_m V_{j}}.
\end{equation}
The nest parameters $\mu_m$ are learnt as scalar constants, as described in \cref{sec:generalisation_framework}.

\paragraph{Cross-nested logit} Cross-nested logit models extend nested logit models by letting an alternative belong to several nests, with a degree of membership $\alpha_{im} \in [0, 1]$ of alternative $i$ to nest $m$. Defining $Y_m = \sum_{j\in m}\alpha_{jm}^{\mu_m}e^{\mu_m V_{j}}$, the choice probability sums over nests:
\begin{equation} \label{eq:cnl_probability}
    P_i = \sum_{m=1}^{M} \frac{\alpha_{im}^{\mu_m}e^{\mu_m V_{i}}}{Y_m}\,\frac{Y_m^{1/\mu_m}}{\sum_{p=1}^{M} Y_p^{1/\mu_p}}.
\end{equation}
The nest parameters $\mu_m$ and degree of membership $\alpha_{im}$ are learnt as scalar constants, as described in \cref{sec:generalisation_framework}.

\paragraph{Ordinal logit} Ordinal logit models are used for ordered categorical outcomes with unknown distance between categories. They model ordered categorical responses for a single alternative, such as Likert-scale questions or star ratings. As such, they compare a single latent utility against thresholds for each ordinal category. Let $y^* = V + \varepsilon$, where $V$ is now the only deterministic utility and $\varepsilon$ follows a standard logistic distribution. Thresholds $\tau_1 < \dots < \tau_{J-1}$ partition the latent continuous space into $J$ ordered categories, and
\begin{equation} \label{eq:ol_binary}
    \Pr(y^* > \tau_j) = \sigma(V - \tau_j) = \frac{1}{1 + e^{-(V - \tau_j)}}, \qquad j = 1, \dots, J-1,
\end{equation}
from which the category probabilities follow:
\begin{equation} \label{eq:ol_probability}
\begin{aligned}
    P_1 &= 1 - \sigma(V - \tau_1), \\
    P_j &= \sigma(V - \tau_{j-1}) - \sigma(V - \tau_j), \qquad j = 2, \dots, J-1, \\
    P_J &= \sigma(V - \tau_{J-1}).
\end{aligned}
\end{equation}
The aggregation operator thus produces one utility per observation; the probability function becomes \cref{eq:ol_probability}, and the loss remains \cref{eq:cel}. Thresholds are learnt as scalar constants, as described in \cref{sec:generalisation_framework}

\subsection{Changing aggregator and probability functions: latent class}\label{sec:ol_lc}

\paragraph{Latent class} Latent class models describe discrete preference heterogeneity: the population consists of classes with different utility functions. With $C$ classes, the modeller specifies utilities $V_{cj}$ for every class~$c$ and alternative~$j$, which may contain different features and parameters. The aggregation function becomes:
\begin{equation}
    V_{cj} = \sum_{k=1}^K \delta_{ck}f_{cjk}(x_k)
\end{equation}
where $\delta_{cj}$ is the Kronecker delta function, indicating whether the coefficient function of variable~$k$ affects the utility function of alternative~$j$ for people in class~$c$.
The probability function becomes a combination
\begin{equation} \label{eq:lc_probability}
    P_i = \sum_{c=1}^C\pi_c P_{ci},
    \qquad
    P_{ci} = \frac{e^{V_{ci}}}{\sum_{j=1}^{J}e^{V_{cj}}},
    \qquad
    \sum_{c=1}^C\pi_c=1
\end{equation}
where $\pi_c$ is the probability of belonging to class~$c$. The loss is given by \cref{eq:cel}. Again, $\pi_i$ are learnt as scalar parameters, as described in \cref{sec:generalisation_framework}.

\subsection{Inter-individual heterogeneity through functional effects: mixed logit and ICLV}\label{sec:heterogeneity}

The linear-in-parameters \acrshort{mxl} model relaxes the i.i.d. assumption of the \acrshort{mnl} by representing unobserved heterogeneity with random terms. \acrshort{iclv} models, by contrast, include unobserved random latent variables anchored using psychometric indicators in the utility function. Both models lack a closed-form probability function, so they must be integrated out for estimation, e.g., by Monte Carlo simulation. Their \acrshort{paramboost} counterparts replace these random terms with deterministic functions of socio-demographic characteristics, learnt as functional effects (\cref{sec:learning_f}). They therefore capture systematic rather than random heterogeneity and are not mixture models in the strict sense; we keep the names \acrshort{mxl} and \acrshort{iclv} to indicate the baseline models they are compared with. The \acrshort{paramboost} versions of these choice models require no simulation because there are no random values to integrate.

\paragraph{Mixed logit} \acrshort{mxl} models include random coefficients $\beta_{il} = \bar{\beta}_{il} + \sigma_{il}\xi$ with $\xi \sim \mathcal{N}(0, 1)$, and
\begin{equation} \label{eq:mxl_utility}
    V_i = \text{ASC}_i + \sum_{k=1}^{K}\beta_{ik}x_{k} + \sum_{l=1}^{L} (\bar{\beta}_{il} + \sigma_{il}\xi )x_l,
\end{equation}
where $k$ indexes features with fixed coefficients and $l$ features with random coefficients. Conditional on $\xi$, the model is an \acrshort{mnl}, and the loss integrates the logit probability over the distribution of $\xi$. Using Monte-Carlo simulation, we have:
\begin{equation} \label{eq:mxl_loss}
    \mathcal{L} = -\frac{1}{N}\sum_{n=1}^{N}\sum_{j=1}^{J} y_{jn}\ln \int P_{jn}(\xi)\,\phi(\xi)\,\mathrm{d}\xi
    \;\approx\; -\frac{1}{N}\sum_{n=1}^{N}\sum_{j=1}^{J} y_{jn}\ln \left( \frac{1}{R} \sum_{r=1}^{R} P_{jn}(\xi_r) \right),
\end{equation}
where the approximation uses $R$ draws $\xi_r$ from $\mathcal{N}(0, 1)$.

In the \acrshort{paramboost} version, the random parameters are functional effects of the socio-demographic characteristics $\mathbf{s}$:
\begin{equation} \label{eq:mxl_fe_utility}
    V_i = \text{ASC}_i + \sum_{k=1}^{K}f_{ik}(x_{k}) + \sum_{l=1}^{L} f_{il}^{\beta}(\mathbf{s})\,x_l,
\end{equation}
where $f_{ik}$ are the piecewise polynomial coefficients of \cref{eq:piecewise_polynomial} and $f_{il}^{\beta}$ the parameter-space ensembles of \cref{eq:functional_effect}. Since \cref{eq:mxl_fe_utility} is deterministic, the probability function and loss are those of the \acrshort{mnl}, and heterogeneity is described by the empirical distribution of $f^{\beta}(\mathbf{s})$ over individuals (\cref{sec:heterogeneity_results}).

\paragraph{Integrated choice and latent variable model} \acrshort{iclv} models include unobserved latent variables in the utility function, derived from a structural equation and anchored using psychometric indicators. Without loss of generalisability, assuming one latent variable, the structural equation links $O$ socio-demographic characteristics to the latent variable:
\begin{equation} \label{eq:structural}
    X^*_n = \sum_{o=1}^{O}\lambda_o s_{on} + \omega_n, \qquad \omega_n \sim \mathcal{N}(0, 1),
\end{equation}
and $Q$ measurement equations anchor it to attitudinal indicators,
\begin{equation} \label{eq:measurement}
    I^*_{qn} = \alpha_q + \delta_q X^*_n + \zeta_{qn}, \qquad \zeta_{qn} \sim \mathcal{N}(0, \sigma_q^2), \qquad q = 1, \dots, Q.
\end{equation}
The indicators are observed on a $T$-point Likert scale, $I_{qn} = t$ if $\tau_{q,t-1} < I^*_{qn} \leq \tau_{q,t}$, with $\tau_{q,0} = -\infty$ and $\tau_{q,T} = \infty$, which gives the ordered probit probabilities
\begin{equation} \label{eq:indicator_probability}
    \Pr(I_{qn} = t \mid X^*_n) = \Phi\left(\frac{\tau_{q,t} - \alpha_q - \delta_q X^*_n}{\sigma_q}\right) - \Phi\left(\frac{\tau_{q,t-1} - \alpha_q - \delta_q X^*_n}{\sigma_q}\right).
\end{equation}
The utility~$j$ becomes $V_{jn} = \text{ASC}_j + \sum_k \beta_{jk}\, x_{kn} + \beta_\text{LV} X^*_n$, and, conditional on $X^*_n$, the choice probability $P(i_n \mid X^*_n)$ is the \acrshort{mnl} probability as in \cref{eq:mnl_probability}. The likelihood of observation $n$ is the joint probability of the choice and the indicators, integrated over $\omega$ and approximated with $R$ draws:
\begin{equation} \label{eq:iclv_likelihood}
    \ell_n = \int P(i_n \mid X^*_n) \prod_{q=1}^{Q}\Pr(I_{qn} \mid X^*_n)\, \phi(\omega)\,\mathrm{d}\omega
    \;\approx\; \frac{1}{R}\sum_{r=1}^{R} P(i_n \mid X^*_{nr}) \prod_{q=1}^{Q}\Pr(I_{qn} \mid X^*_{nr}),
\end{equation}
where $X^*_{nr}$ is obtained with \cref{eq:structural} evaluated at draw $\omega_r$. The loss is $\mathcal{L} = -\frac{1}{N}\sum_n \ln \ell_n$.

In the \acrshort{paramboost} version, the structural equation, \cref{eq:structural}, is replaced by a utility-space ensemble:
\begin{equation} \label{eq:latent_variable_fe}
    X^*_n = f^{V}(\mathbf{s}_n),
\end{equation}
such that the latent variable is deterministic. The measurement equations, \cref{eq:measurement}--\cref{eq:indicator_probability} still hold, and anchor the latent variable. Because there is no simulation, the likelihood reduces to $\ell_n = ln(P(i_n \mid X^*_n)\prod_q \Pr(I_{qn} \mid X^*_n)$). Therefore, the gradients used to learn the latent variable are composed of two summed parts:
\begin{inline}
    \item the loss on the choices; and
    \item the loss on the $Q$ measurement equations.
\end{inline}
Finally, since the ensemble output is already in the utility space, we don't estimate $\beta^{LV}$ for ParamBoost.

%% ===================================================================
\section{Empirical evaluation}\label{sec:experiments}

\subsection{Experimental setup}\label{sec:setup}

\paragraph{Data} We use the Swissmetro \citep{bierlaire2001acceptance} and Optima \citep{bierlaire2018mode} datasets and adapt the examples given on the Biogeme \citep{bierlaire2023short} website\footnote{https://biogeme.epfl.ch/sphinx/examples.html} to the different choice models. \Cref{tab:datasets} lists the features and dataset summary statistics. We use the Swissmetro dataset for all models except \acrshort{iclv}, which requires attitudinal indicators available in the Optima dataset. 
\begin{table}[htbp]
\centering
\caption{Datasets and features.}
\label{tab:datasets}
\footnotesize
\begin{tabular}{lll}
\toprule
 & Swissmetro & Optima \\
\midrule
Trip attributes & \makecell[l]{Travel time\\ Cost} & \makecell[l]{Travel time\\ Cost\\ Distance} \\
\addlinespace
Socio-demographic characteristics & \makecell[l]{Gender\\ First class seat\\ Purpose of the trip\\ Age category} & \makecell[l]{Manager\\ Car-oriented parents\\ High education degree\\ Low education degree\\ Used to go to school by car} \\
\addlinespace
Observations & 9036 & 889 \\
\addlinespace
Alternatives & Train, Swissmetro, car & Car, public transport, slow modes \\
\bottomrule
\end{tabular}
\end{table}

Two models use modified versions of Swissmetro: 
\begin{inline}
    \item for the \acrshort{ol} model, the three alternatives are treated as ordered categories, purely for illustration, using only the train cost and travel time as input features; and
    \item for the \acrshort{bp} model, we remove observations choosing the Swissmetro alternative to obtain a binary choice between train and car.
\end{inline}
\Cref{tab:modelling-decisions} summarises the key modelling decisions for all models.

\begin{table}[htbp]
\centering
\small
\caption{Summary of key modelling decisions for each model. Unless stated otherwise, every model uses the same utility structure as the \acrshort{mnl}.}
\label{tab:modelling-decisions}
\renewcommand{\arraystretch}{1.15}
\begin{tabularx}{\linewidth}{@{}l >{\raggedright\arraybackslash}p{0.28\linewidth} >{\raggedright\arraybackslash}X@{}}
\toprule
Model & Full name & Key modelling decisions \\
\midrule
\acrshort{mnl}  & \acrlong{mnl}  & Alternative-specific parameters for travel time and cost features with ASCs (Swissmetro ASC normalised to 0) \\
\acrshort{dl}   & \acrlong{dl}   & Previous choice omitted when missing. Swissmetro choice normalised to 0 \\
\acrshort{bp}   & \acrlong{bp}   & Binary dataset with train and car alternatives only \\
\acrshort{nl}   & \acrlong{nl}   & Nest for train and car (existing alternatives) \\
\acrshort{cnl}  & \acrlong{cnl}  & Nest for train and car (existing alternatives) and for train and Swissmetro (public transport) \\
\acrshort{ol}   & \acrlong{ol}   & Only using train travel time and cost for illustration \\
\acrshort{lc}   & \acrlong{lc}   & Only cost in class 1 utility, cost and travel time in class 2 utility \\
\acrshort{mxl}  & \acrlong{mxl}  & Travel time parameters are random \\
\acrshort{iclv} & \acrlong{iclv} & Car loving latent variable in the car utility \\
\bottomrule
\end{tabularx}
\end{table}

\paragraph{Features and constraints} All trip \acrshort{los} attributes are constrained to decrease utility and, where applicable, to be convex, so that a given increase in an attribute has a stronger effect at low values than at high values. However, we do not apply these constraints for the \acrshort{ol} model, since the ordering of its response is artificial and domain knowledge is irrelevant in this case. We use socio-demographic characteristics only in the \acrshort{iclv} models and in the \acrshort{paramboost} \acrshort{mxl} models.

\paragraph{Models and estimation} For every choice model, we compare the four \acrshort{paramboost} variants in bold in \cref{tab:piecewise_poly} (ParamB-$O^0$-$C^{-1}$, ParamB-$O^1$-$C^0$, ParamB-$O^3$-$C^-1$, and ParamB-$O^3-C^2$) with a linear-in-parameters baseline estimated with Biogeme \citep{bierlaire2023short}\footnote{Note, the framework supports estimation of linear-in-parameters models, but we rely on Biogeme to ensure independence of the benchmark.}. Baselines that require simulation (\acrshort{mxl} and \acrshort{iclv}) use 500 draws generated with modified Latin hypercube sampling \citep{hess2006use}. For the \acrshort{paramboost} models, we select the number of boosting iterations by 5-fold cross-validation, with at most 3000 iterations and early stopping after 100 iterations without improvement. All other hyperparameters are set to their default values (\cref{app:hyper}).

\paragraph{Evaluation} We split each dataset into a training set with 70\% of the observations and a test set with the remaining 30\%. We split at the individual level, so repeated choices of the same individual never appear in both sets. We report the mean \acrshort{cel} (\cref{eq:cel}) on the test set. This means we do not report the contribution of the measurement equations for the \acrshort{iclv}.

\subsection{Predictive performance}\label{sec:performance}

\Cref{tab:results} reports the test \acrshort{cel} of all models; training and test values are given in \cref{app:cel}. We observe four main findings.

\begin{table}[htbp]
\centering
\caption{Test \acrshort{cel} (lower is better). The best value for each model is in bold; $\Delta$ is the improvement of the best \acrshort{paramboost} variant over the linear-in-parameters baseline.}
\label{tab:results}
\footnotesize
\begin{tabular}{lrrrrrr}
\toprule
Model & \pb{0}{-1} & \pb{1}{0} & \pb{3}{-1} & \pb{3}{2} & Linear & $\Delta$ \\
\midrule
\multicolumn{7}{l}{\textit{Multi-class classification on the Swissmetro dataset}} \\ \midrule
\acrshort{mnl}  & 0.7911 & 0.7924 & \textbf{0.7909} & 0.7924 & 0.8119 & 0.0210 \\
\acrshort{dl}   & 0.7227 & 0.7267 & \textbf{0.7221} & 0.7283 & 0.7448 & 0.0227 \\
\acrshort{nl}   & 0.7950 & 0.7961 & \textbf{0.7946} & 0.7962 & 0.8120 & 0.0174 \\
\acrshort{cnl}  & 0.7925 & 0.7936 & \textbf{0.7923} & 0.7938 & 0.8128 & 0.0205 \\
\acrshort{lc}   & \textbf{0.7875} & 0.7942 & 0.8354 & 0.7956 & 0.8107 & 0.0232 \\
\acrshort{mxl}  & \textbf{0.8007} & 0.8132 & 0.8049 & 0.8110 & 0.8206 & 0.0199 \\ \midrule
\multicolumn{7}{l}{\textit{Binary classification on the Swissmetro dataset}} \\ \midrule
\acrshort{bp}   & 0.3627 & 0.3652 & \textbf{0.3607} & 0.3787 & 0.4676 & 0.1069 \\ \midrule
\multicolumn{7}{l}{\textit{Ordinal classification on the Swissmetro dataset}} \\ \midrule
\acrshort{ol}   & 0.8536 & 0.8477 & 0.8540 & \textbf{0.8432} & 0.8627 & 0.0195 \\ \midrule
\multicolumn{7}{l}{\textit{Multi-class classification on the Optima dataset}} \\ \midrule
\acrshort{iclv} & 0.5005 & \textbf{0.4822} & 0.4902 & 0.5047 & 0.5588 & 0.0766 \\
\bottomrule
\end{tabular}
\end{table}

First, across comparable model structures (i.e., a multi-class classification task on the Swissmetro dataset), the best model structures are \acrshort{dl} and \acrshort{lc}. While the \acrshort{dl} model structure incorporates more information from previous choices, the \acrshort{lc} model structure uses the same amount of information as the \acrshort{mnl}. This is a solid motivation for alternative model structures not yet implemented in most \acrshort{ml} libraries. The \acrshort{lc} model is also the only one that improves on the \acrshort{mnl} for the linear-in-parameters baseline, so the performance is consistent with linear-in-parameter variations. This means that where linear-in-parameter model structures improve predictive performance, the \acrshort{paramboost} counterpart also improves performance compared to a classical \acrshort{mnl}. See, e.g. \cite{salvade2025rumboost, salvade2025functional} for examples where the \acrshort{nl} and \acrshort{mxl} model structures improve on the \acrshort{mnl} model structure using the framework.

Second, boosted utilities consistently improve prediction. In 35 of the 36 comparisons between a framework variant and its linear-in-parameters baseline, the \acrshort{paramboost} model has the lower test \acrshort{cel}. For the \acrshort{iclv}, the best variant improves on the baseline by 0.077; for the \acrshort{bp}, the improvement is 0.107; and for the seven other Swissmetro models, the improvement ranges from 0.017 to 0.023. \pb{3}{-1} in the \acrshort{lc} model is the only exception, as it does not improve over the baseline. Since this variant is the least restricted \acrshort{paramboost} variant, it suffers from numerical instability during training. It converges to a class share far from those of the other models (\cref{app:structural}). In contrast, the \pb{0}{-1} \acrshort{lc} model is numerically stable and outperforms the \acrshort{mnl} model.

Third, the degree and continuity of the utility matter for some models but not as much for others. For the \acrshort{mnl}, \acrshort{nl} and \acrshort{cnl}, the four variants lie within 0.002 of each other, and within 0.006 for the \acrshort{dl}; on a single train--test split, these differences are too small to pick a variant as the best one confidently. For the \acrshort{ol}, \acrshort{mxl}, \acrshort{bp}, \acrshort{iclv} and \acrshort{lc} models the spread is larger, from 0.011 to 0.048, but no variant dominates: the smoothest, \pb{3}{2}, is best for the \acrshort{ol} but worst for the \acrshort{bp}; \pb{1}{0} is best for the \acrshort{iclv}; and \pb{0}{-1} is best for the \acrshort{mxl} and \acrshort{lc}. Smooth utilities help most where: 
\begin{inline}
    \item there are no shape or monotonicity constraints, as in the \acrshort{ol} model; and
    \item the dataset is smaller, as in the Optima dataset for the \acrshort{iclv} model.
\end{inline} 

Finally, smoother utilities overfit less. As seen in \cref{tab:results_full} in \cref{app:cel}, among the \acrshort{paramboost} variants, \pb{3}{2} has the smallest gap between training and test \acrshort{cel} in seven of the nine models (all except the \acrshort{lc} and \acrshort{bp}), and in seven of the nine, the two discontinuous variants have the largest gaps. In the \acrshort{mnl}, for instance, the gap is 0.040 for \pb{3}{2} against 0.066 for both discontinuous variants. On Optima, the test \acrshort{cel} is lower than the training \acrshort{cel} for \pb{1}{0}, \pb{3}{2} and the linear \acrshort{iclv}, which more likely reflects the small test set (about 270 observations) than a systematic effect.

\subsection{Behavioural interpretation}\label{sec:interpretation}

We provide an in-depth analysis of the parameters for \acrshort{mnl}, \acrshort{dl}, \acrshort{mxl}, \acrshort{ol}, and \acrshort{iclv} in the main body of this paper. Other parameters appear in \cref{app:figures}, while structural parameters appear in \cref{app:structural} for conciseness.

\subsubsection{Learnt attribute effects}\label{sec:attribute_effects}

\begin{figure}[htbp]
    \centering
    \begin{subfigure}[b]{0.48\textwidth}
        \centering
        \includegraphics[width=\textwidth]{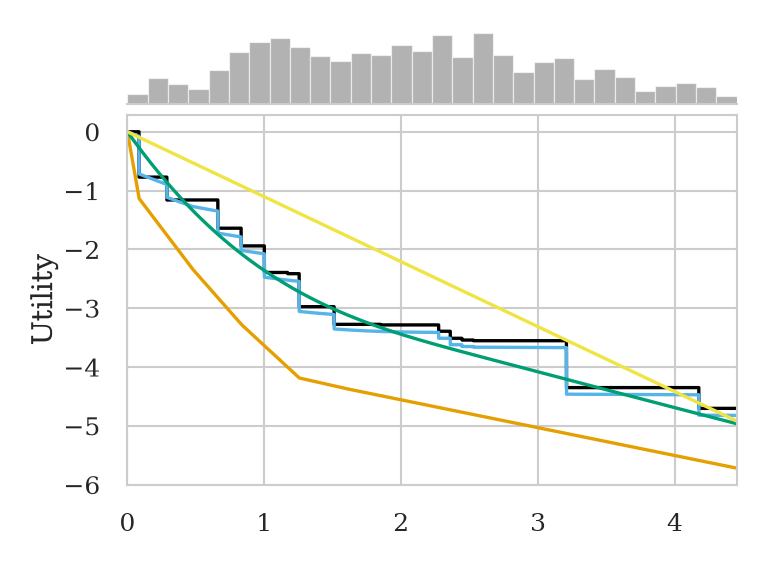}
        \caption{Train: travel time (hours)}
        \label{fig:mnl_train_tt}
    \end{subfigure}
    \hfill
    \begin{subfigure}[b]{0.48\textwidth}
        \centering
        \includegraphics[width=\textwidth]{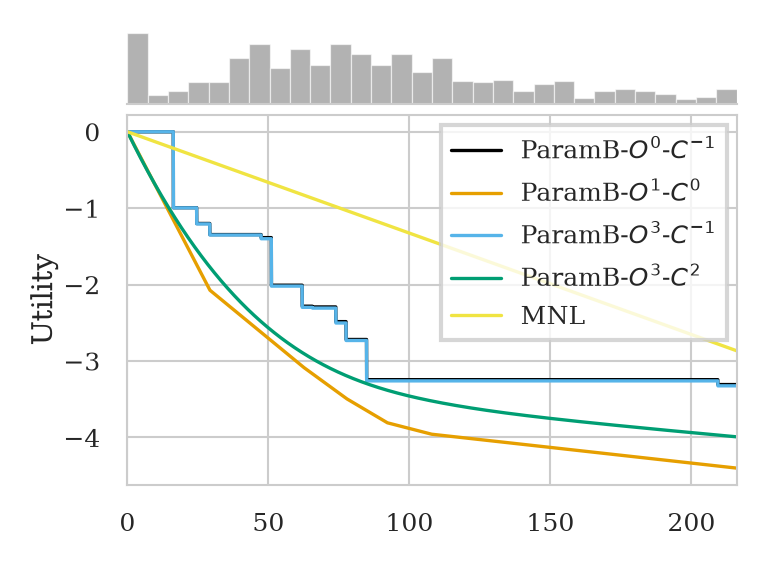}
        \caption{Train: cost (CHF)}
        \label{fig:mnl_train_cost}
    \end{subfigure}

    \vspace{1em}
    \begin{subfigure}[b]{0.48\textwidth}
        \centering
        \includegraphics[width=\textwidth]{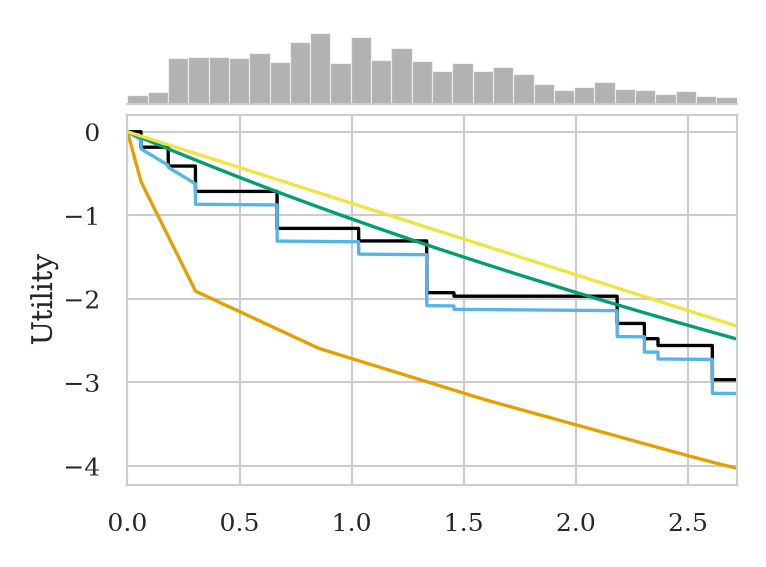}
        \caption{Swissmetro: travel time (hours)}
        \label{fig:mnl_sm_tt}
    \end{subfigure}
    \hfill
    \begin{subfigure}[b]{0.48\textwidth}
        \centering
        \includegraphics[width=\textwidth]{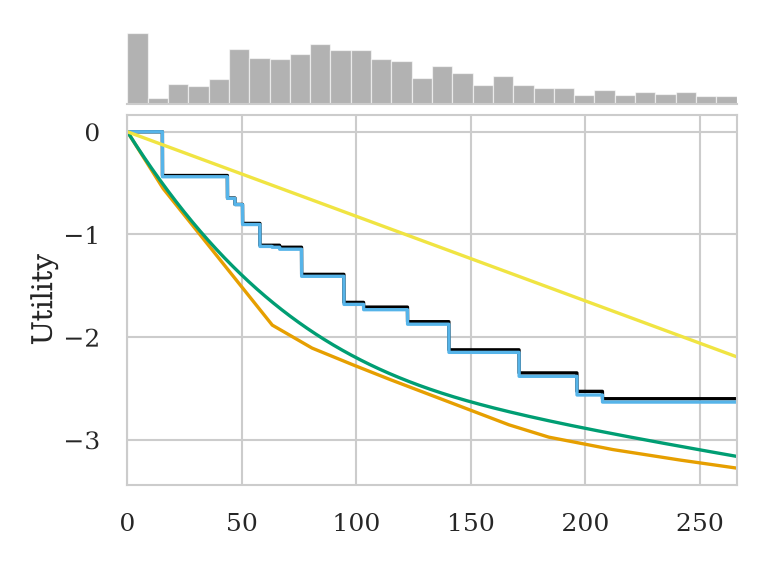}
        \caption{Swissmetro: cost (CHF)}
        \label{fig:mnl_sm_cost}
    \end{subfigure}

    \vspace{1em}
    \begin{subfigure}[b]{0.48\textwidth}
        \centering
        \includegraphics[width=\textwidth]{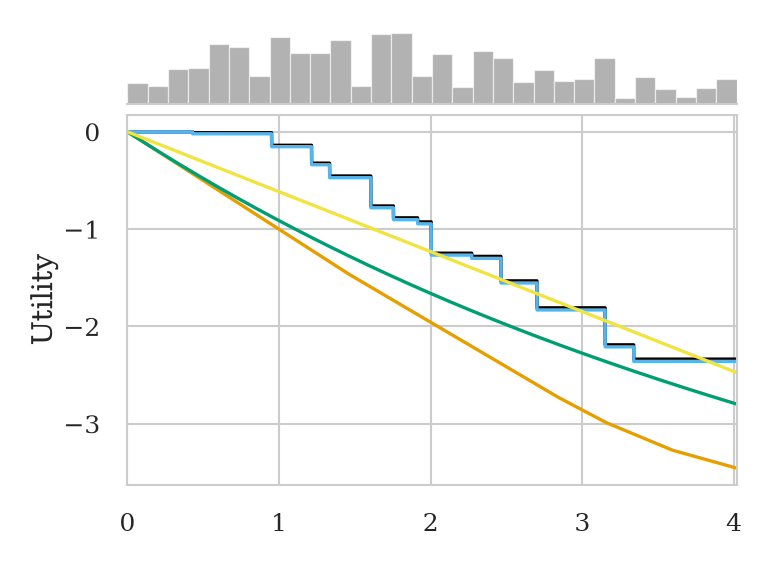}
        \caption{Car: travel time (hours)}
        \label{fig:mnl_car_tt}
    \end{subfigure}
    \hfill
    \begin{subfigure}[b]{0.48\textwidth}
        \centering
        \includegraphics[width=\textwidth]{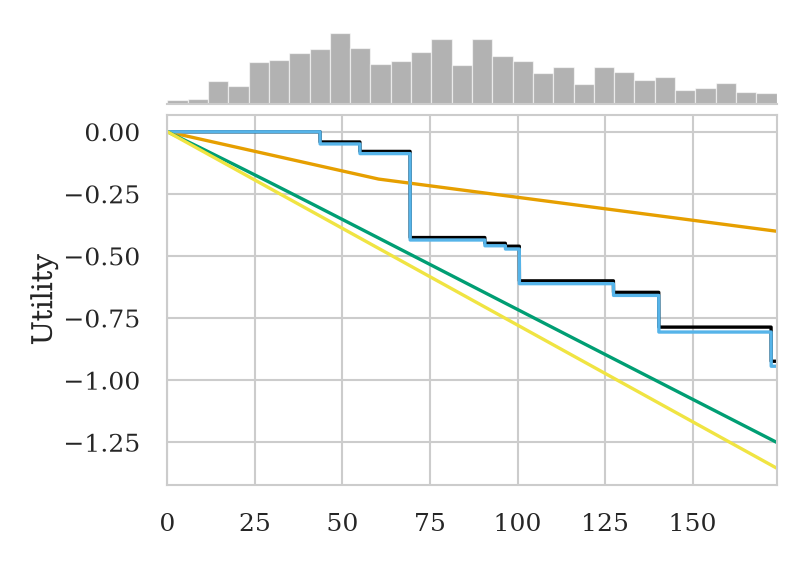}
        \caption{Car: cost (CHF)}
        \label{fig:mnl_car_cost}
    \end{subfigure}

    \caption{Utility contributions of the trip attributes in the \acrshort{mnl}, by alternative and model. Travel time is in hours and cost in CHF. Figures are cropped at the \nth{95} percentile of the feature, while its distribution is shown above the x-axis.}
    \label{fig:mnl_utilities}
\end{figure}

\Cref{fig:mnl_utilities} shows the learnt utility contributions of travel time and cost in the \acrshort{mnl}. In regions with many observations, the four \acrshort{paramboost} variants behave similarly. They differ at extreme values, where data are sparse, and the continuity constraints force smoother extrapolation. Combined with the shape and monotonicity constraints, the non-linear utilities exhibit diminishing marginal disutility. In contrast, the linear-in-parameters \acrshort{mnl} must extrapolate linearly and assigns very large disutilities to long or expensive trips. The \acrshort{nl}, \acrshort{cnl}, \acrshort{dl}, \acrshort{lc} and \acrshort{bp} models show the same pattern and are shown in \cref{app:figures}. All utility functions respect the monotonicity and shape constraints.

\begin{figure}[htbp]
    \centering
    \begin{subfigure}[b]{0.48\textwidth}
        \centering
        \includegraphics[width=\textwidth]{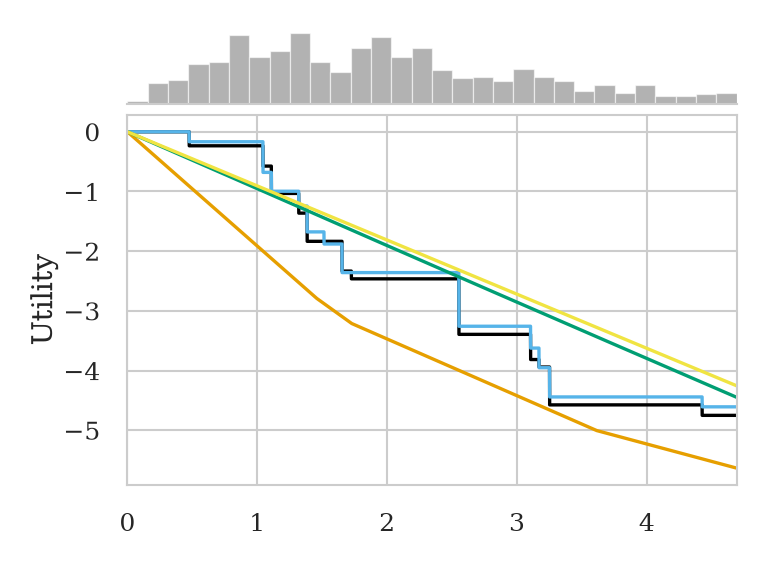}
        \caption{Public transport: travel time (hours)}
        \label{fig:iclv_pt_tt}
    \end{subfigure}
    \hfill
    \begin{subfigure}[b]{0.48\textwidth}
        \centering
        \includegraphics[width=\textwidth]{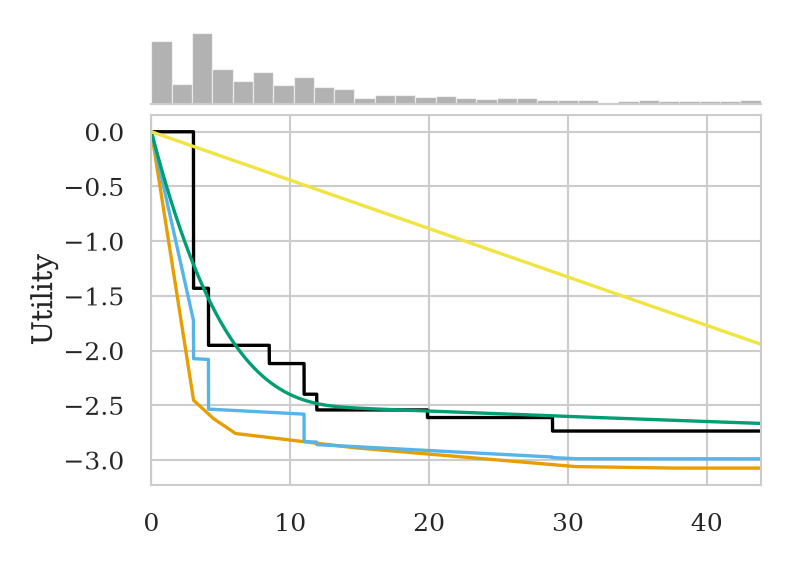}
        \caption{Public transport: cost (CHF)}
        \label{fig:iclv_pt_cost}
    \end{subfigure}

    \vspace{1em}
    \begin{subfigure}[b]{0.48\textwidth}
        \centering
        \includegraphics[width=\textwidth]{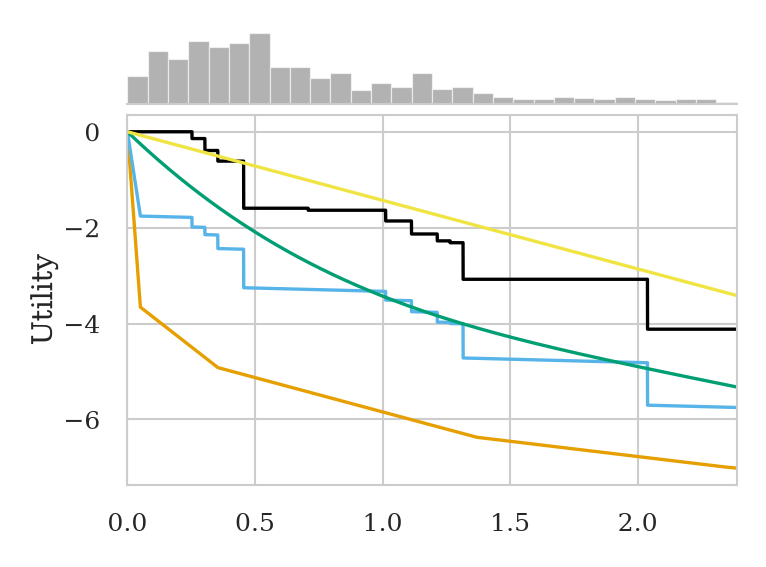}
        \caption{Car: travel time (hours)}
        \label{fig:iclv_car_tt}
    \end{subfigure}
    \hfill
    \begin{subfigure}[b]{0.48\textwidth}
        \centering
        \includegraphics[width=\textwidth]{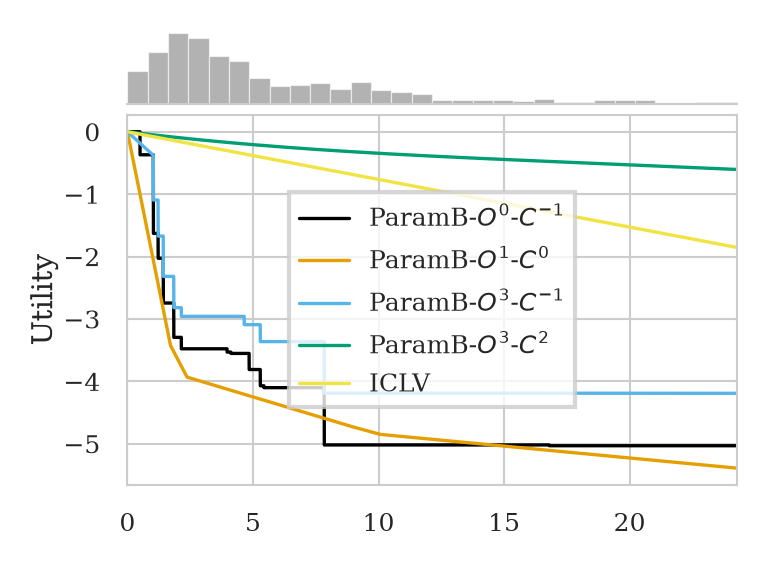}
        \caption{Car: cost (CHF)}
        \label{fig:iclv_car_cost}
    \end{subfigure}

    \vspace{1em}
    \begin{subfigure}[b]{0.48\textwidth}
        \centering
        \includegraphics[width=\textwidth]{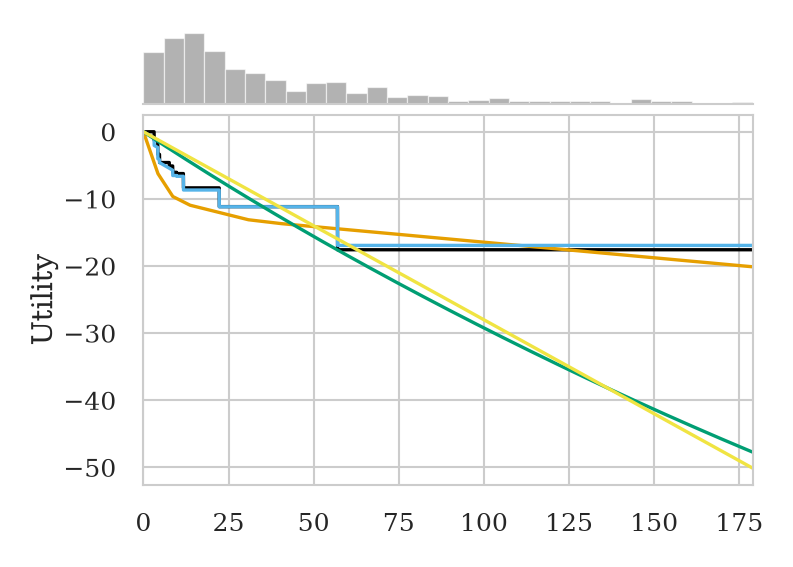}
        \caption{Slow modes: distance (Km)}
        \label{fig:iclv_distance}
    \end{subfigure}
    \hfill
    \begin{subfigure}[b]{0.48\textwidth}
        \centering
        \includegraphics[width=\textwidth]{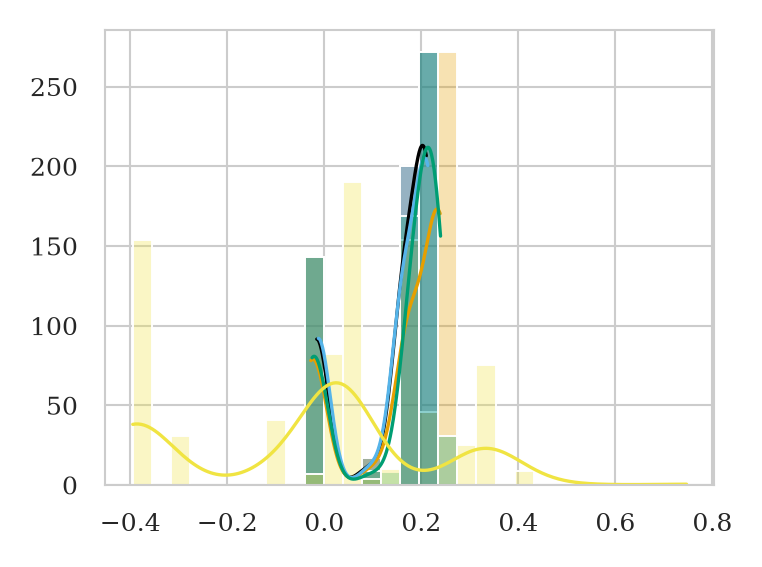}
        \caption{Latent variable: car-loving attitude}
        \label{fig:iclv_lv}
    \end{subfigure}

    \caption{Utility contributions of the trip attributes in the \acrshort{iclv} (Optima), by model, and distribution of the latent variable (panel f). Travel time is in hours, cost in CHF and distance in km. Figures are cropped at the \nth{95} percentile of the feature, while its distribution is shown above the x-axis.}
    \label{fig:iclv_utilities}
\end{figure}

On Optima (\cref{fig:iclv_utilities}), travel times for car and public transport exhibit relatively low non-linearity for all \acrshort{paramboost} variants, except for \pb{1}{0}. Costs, in contrast, show a sharp initial decrease followed by a plateau for most variants, unlike the linear \acrshort{iclv}. The distance effect for slow modes is almost linear for \pb{3}{2}, close to the linear \acrshort{iclv}, whereas the other variants flatten or increase more slowly beyond 100km. This difference may contribute to \pb{3}{2} 's weaker test performance in \acrshort{iclv}.

\subsubsection{Utilities without shape constraints}\label{sec:unconstrained}

\begin{figure}[htbp]
    \centering
    \begin{subfigure}[b]{0.48\textwidth}
        \centering
        \includegraphics[width=\textwidth]{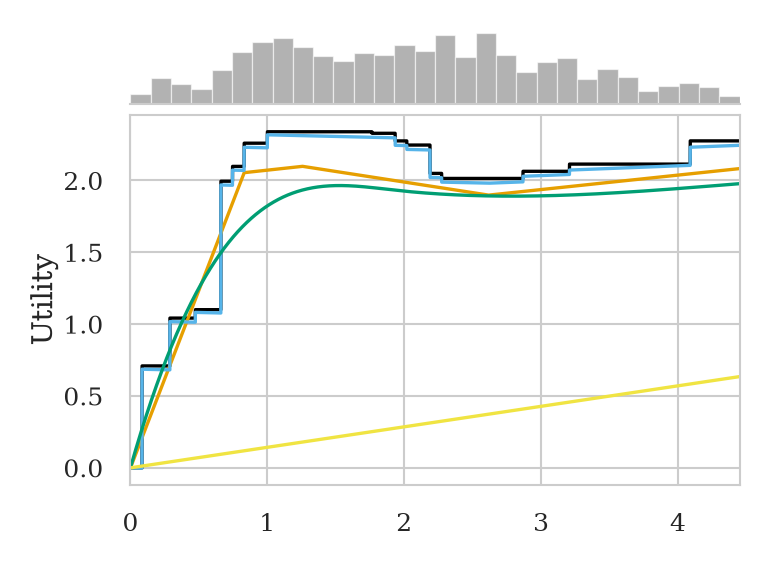}
        \caption{Train: travel time (hours)}
        \label{fig:ol_tt}
    \end{subfigure}
    \hfill
    \begin{subfigure}[b]{0.48\textwidth}
        \centering
        \includegraphics[width=\textwidth]{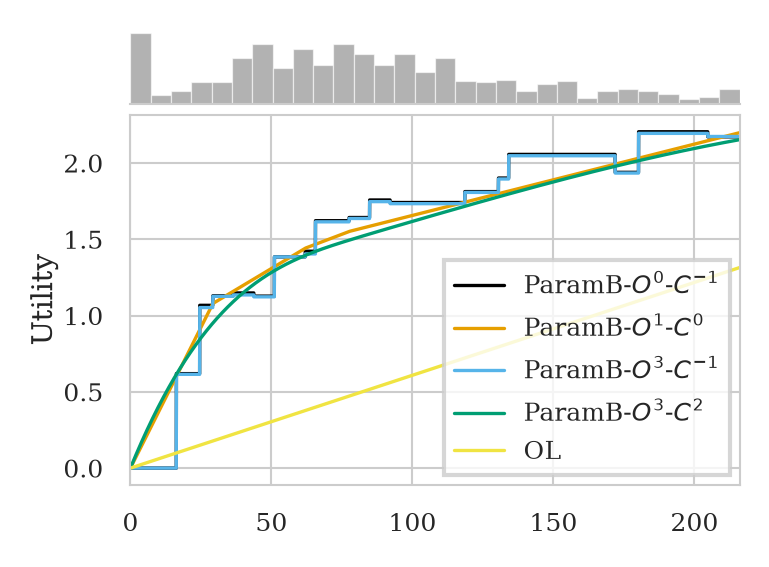}
        \caption{Train: cost (CHF)}
        \label{fig:ol_cost}
    \end{subfigure}
    \caption{Utility contributions in the \acrshort{ol}, estimated without shape constraints. Travel time is in hours and cost in CHF. Figures are cropped at the \nth{95} percentile of the feature, while its distribution is shown above the x-axis.}
    \label{fig:ol_utilities}
\end{figure}

The \acrshort{ol} model is estimated without monotonicity or convexity constraints, i.e. without regularisation. \Cref{fig:ol_utilities} shows that, without them, the discontinuous variants can fit the training data too closely, while the smooth variants recover more regular effects. This is consistent with the second and third findings of \cref{sec:performance}: smoothness matters most when no shape constraints are imposed, and in that setting \pb{3}{2} generalises best. The figure also shows the flexibility that the non-linear models gain over the linear-in-parameters \acrshort{ol}.

\subsubsection{Preference heterogeneity}\label{sec:heterogeneity_results}

\begin{figure}[htbp]
    \centering
    \begin{subfigure}[b]{0.48\textwidth}
        \centering
        \includegraphics[width=\textwidth]{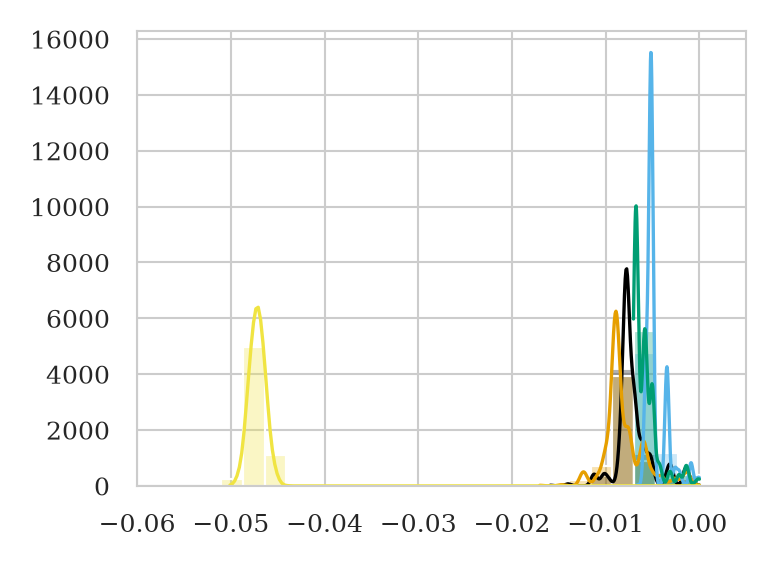}
        \caption{Train: travel time (1/hours)}
        \label{fig:mxl_train_tt}
    \end{subfigure}
    \hfill
    \begin{subfigure}[b]{0.48\textwidth}
        \centering
        \includegraphics[width=\textwidth]{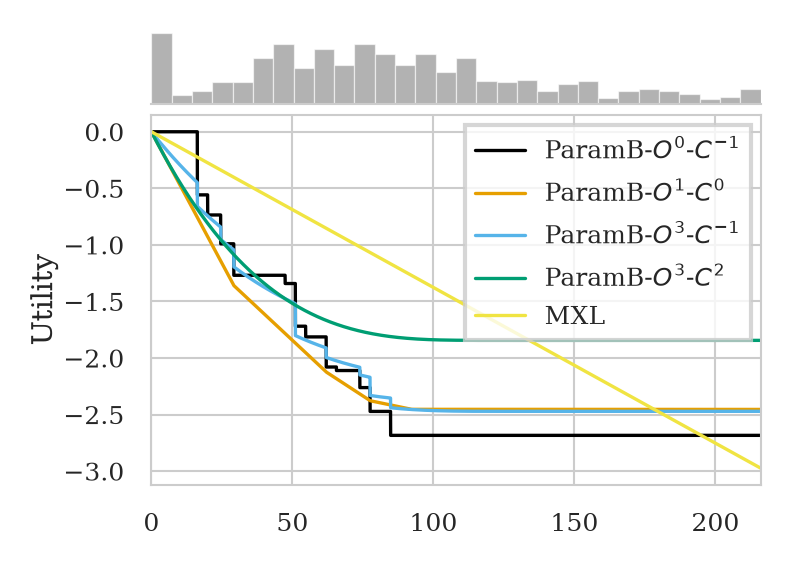}
        \caption{Train: cost (CHF)}
        \label{fig:mxl_train_cost}
    \end{subfigure}

    \vspace{1em}
    \begin{subfigure}[b]{0.48\textwidth}
        \centering
        \includegraphics[width=\textwidth]{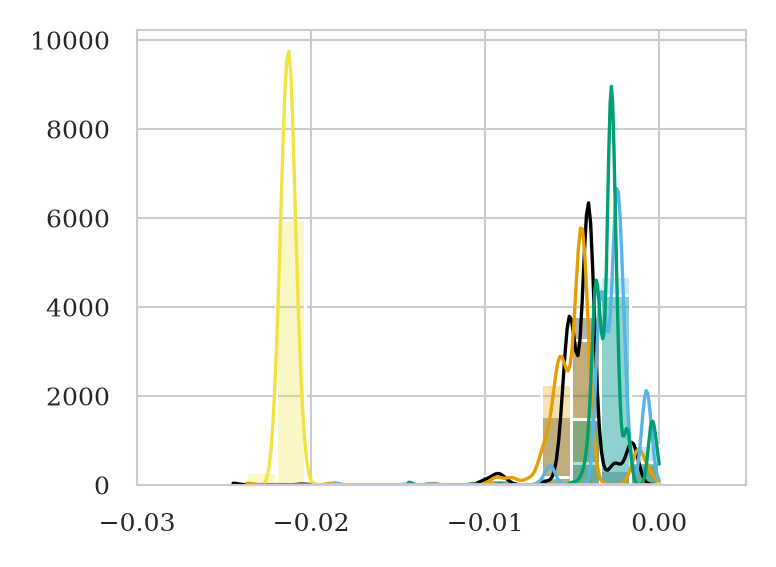}
        \caption{Swissmetro: travel time (1/hours)}
        \label{fig:mxl_sm_tt}
    \end{subfigure}
    \hfill
    \begin{subfigure}[b]{0.48\textwidth}
        \centering
        \includegraphics[width=\textwidth]{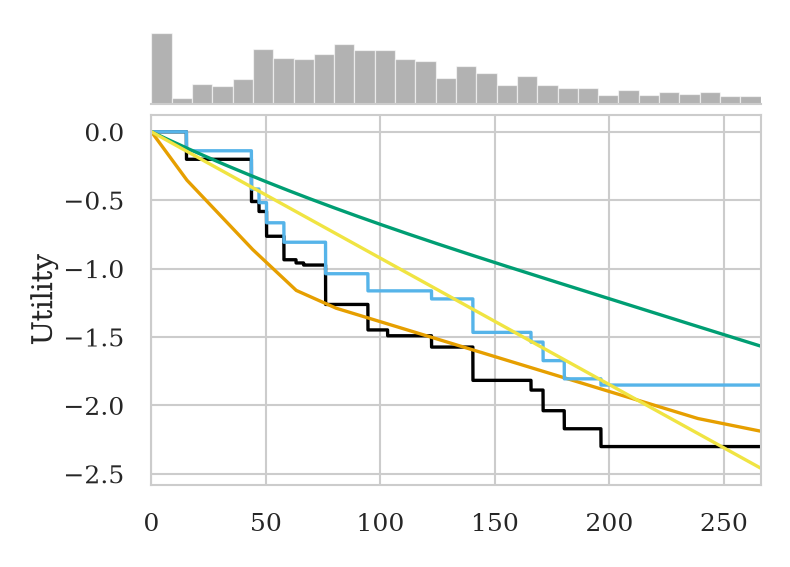}
        \caption{Swissmetro: cost (CHF)}
        \label{fig:mxl_sm_cost}
    \end{subfigure}

    \vspace{1em}
    \begin{subfigure}[b]{0.48\textwidth}
        \centering
        \includegraphics[width=\textwidth]{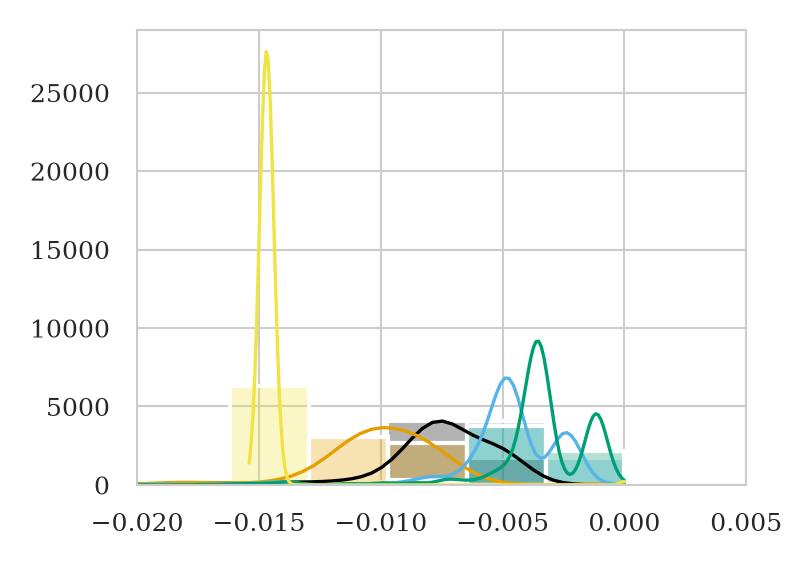}
        \caption{Car: travel time (1/hours)}
        \label{fig:mxl_car_tt}
    \end{subfigure}
    \hfill
    \begin{subfigure}[b]{0.48\textwidth}
        \centering
        \includegraphics[width=\textwidth]{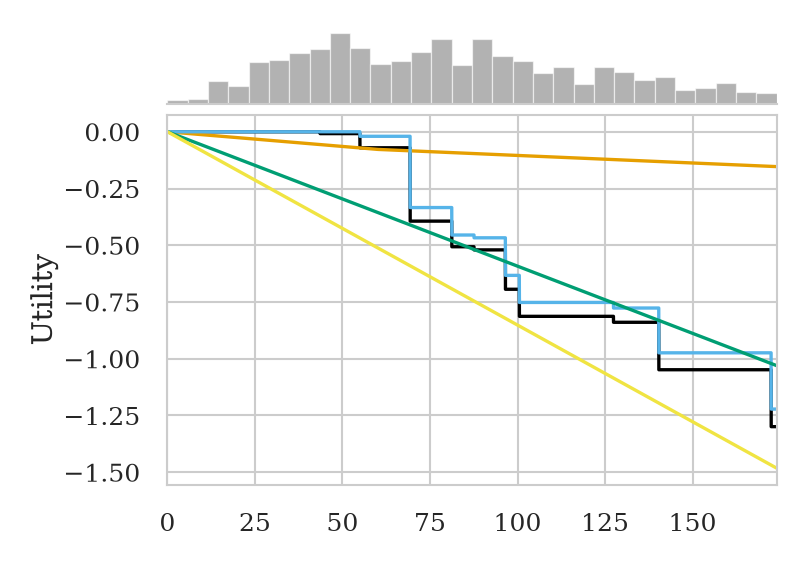}
        \caption{Car: cost (CHF)}
        \label{fig:mxl_car_cost}
    \end{subfigure}

    \caption{\acrshort{mxl}: population distributions of the travel-time parameters (1/hour; a, c, e) and utility contributions of cost (CHF; b, d, f), by model. Figures are cropped at the \nth{95} percentile of the feature, while its distribution is shown above the x-axis.}
    \label{fig:mxl_utilities}
\end{figure}

\Cref{fig:mxl_utilities} shows, for the \acrshort{mxl}, the population distributions of the travel-time parameters and the utility contributions of cost. The cost effects are very similar to those of the \acrshort{mnl}. The baseline \acrshort{mxl} has a more negative mean travel-time parameter than all \acrshort{paramboost} variants. The \acrshort{paramboost} distributions are roughly bell-shaped, and the variants with a higher maximum degree have slightly less negative means. All travel-time parameters are negative, as required by the monotonicity constraint (\cref{sec:learning_f}).

The latent variable of the \acrshort{iclv} (\cref{fig:iclv_lv}) is similar across \acrshort{paramboost} variants and bimodal: one group of individuals has a car-loving attitude that increases the utility of car, and a second, neutral or car-averse group has an attitude that reduces it slightly. The baseline latent variable is more dispersed, with roughly three modes that can be read as dislike of, neutrality towards and liking of cars.

\subsubsection{State dependence}\label{sec:state_dependence}

\Cref{fig:dl_utilities} shows the effect of the previous choice in the \acrshort{dl}, with Swissmetro as the reference. Choosing the car or train alternative in the previous choice situation substantially increases car and train utility, respectively, compared with choosing the Swissmetro alternative in previous choice situations.
Since Swissmetro is a stated-preference survey, this reflects inertia across the choice tasks of the same respondent rather than habit formation in real travel. It nevertheless shows that repeated choices are correlated and should be modelled accordingly.

\begin{figure}[htbp]
    \centering
    \begin{subfigure}[b]{0.48\textwidth}
        \centering
        \includegraphics[width=\textwidth]{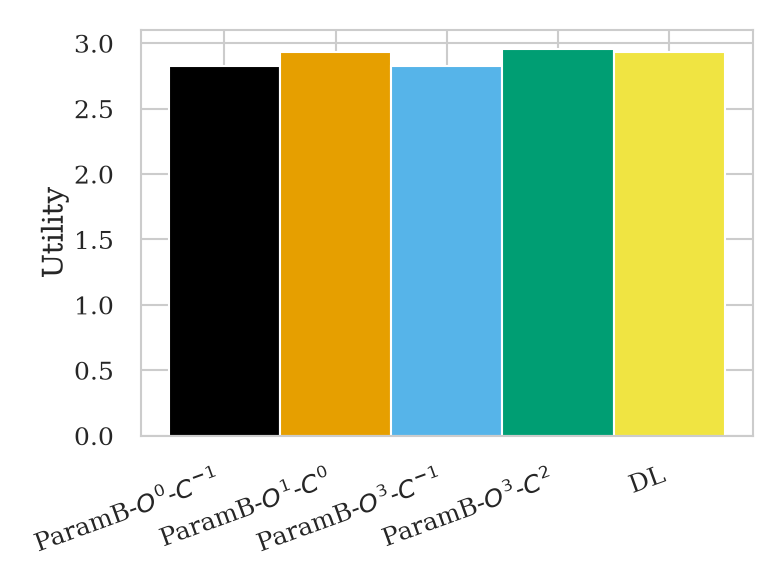}
        \caption{Train chosen in previous choice situation}
        \label{fig:dl_train}
    \end{subfigure}
    \hfill
    \begin{subfigure}[b]{0.48\textwidth}
        \centering
        \includegraphics[width=\textwidth]{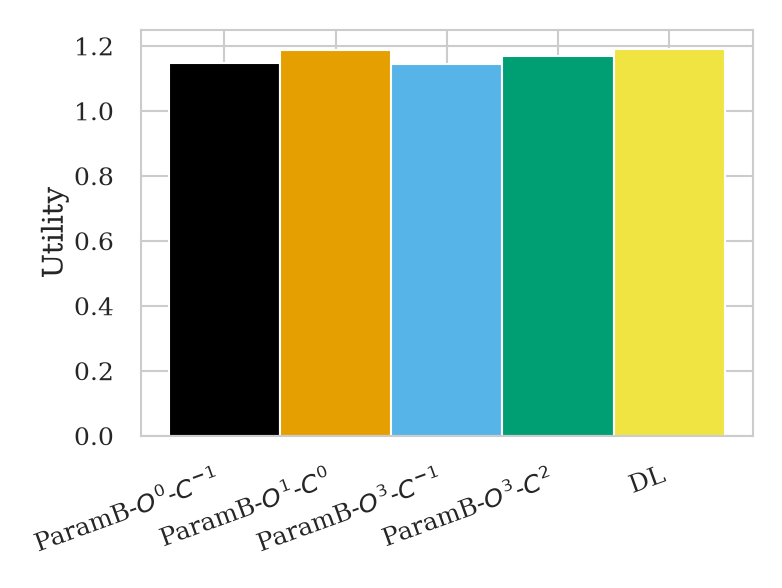}
        \caption{Car chosen in previous choice situation}
        \label{fig:dl_car}
    \end{subfigure}
    \caption{Utility contribution of the previous choice in the \acrshort{dl}, by model.}
    \label{fig:dl_utilities}
\end{figure}

%% ===================================================================
\section{Conclusion}\label{sec:conclusion}

In this paper, we have introduced a unified interpretable non-linear choice modelling framework based on gradient boosting coefficient functions. We showed how to implement nine different models within this framework and demonstrated how easily it can be adapted and extended to other model formulations. We have also demonstrated the value added by the non-linearities by benchmarking against a linear-in-parameters baseline. We also demonstrated the value of complex model structures by showing that the \acrshort{lc} model achieves best-in-class performance on the Swissmetro case study. In addition, we show a trade-off between constraints, interpretability, and predictive performance, and that continuity/smoothness constraints act as a regulariser. 

Whilst the experiments are intended to demonstrate the ability to implement arbitrary model specifications and their benefit compared to linear-in-parameters models, we have restricted them to simple datasets, sometimes with hypothetical semi-synthetic model structures. Demonstrating the framework's true value in systematically improving performance will require evaluation on larger, more complex datasets, particularly those where alternative model specifications clearly outperform MNL in traditional linear-in-parameters models. That said, we have shown in prior work \citep{salvade2025rumboost, salvade2025functional} that alternative model structures such as the \acrshort{nl}, \acrshort{ol}, and \acrshort{mxl} clearly improve model performance on large-scale datasets. 

\section*{CRediT authorship contribution statement}

\textbf{Nicolas Salvadé}: Writing -- original draft, Visualization, Software, Methodology, Conceptualization, Formal analysis, Investigation. \textbf{Tim Hillel}: Writing -- review \& editing, Supervision, Methodology, Conceptualization, Project administration.

\section*{Declaration of competing interest}
The authors declare that they have no known competing financial interests or personal relationships that could have appeared to influence the work reported in this paper.

\section*{Acknowledgments}
This research is supported by a Chadwick PhD Scholarship awarded to Nicolas Salvadé by the UCL Department of Civil, Environmental, and Geomatic Engineering.

%% ===================================================================
\appendix
\setcounter{figure}{0}
\setcounter{table}{0}

\FloatBarrier
\section{Hyperparameters}\label{app:hyper}

\begin{table}[htbp]
\centering
\caption{Default \acrshort{paramboost} hyperparameters used for all models. The number of boosting iterations is selected by 5-fold cross-validation (maximum 3000, early stopping after 100 iterations).}
\label{tab:hyperparameters}
\footnotesize
\begin{tabular}{lr}
\toprule
Hyperparameter & Value \\
\midrule
Learning rate & 0.1 \\
Minimum gain to split & 0 \\
Minimum sum of second derivatives in a leaf & 0 \\
Maximum number of histogram bins, degree-0 terms & 64 \\
Maximum number of histogram bins, degree-1 to degree-3 terms & 16 \\
Minimum number of observations in a bin & 3 \\
Minimum number of observations in a leaf & 20 \\
L1 regularisation & $10^{-4}$ \\
L2 regularisation & $10^{-4}$ \\
Maximum number of leaves & 5 \\
Maximum tree depth & None \\
Subsampling fraction & 1 \\
Subsampling frequency & 0 \\
\bottomrule
\end{tabular}
\end{table}

\FloatBarrier
\section{Training and test cross-entropy loss}\label{app:cel}

\begin{table}[htbp]
\centering
\caption{Training and test \acrshort{cel} for all models (lower is better).}
\label{tab:results_full}
\footnotesize
\begin{tabular}{llrrrrr}
\toprule
Model & Set & \pb{0}{-1} & \pb{1}{0} & \pb{3}{-1} & \pb{3}{2} & Linear \\
\midrule
\multicolumn{7}{l}{\textit{Multi-class classification on the Swissmetro dataset}} \\ \midrule
\multirow{2}{*}{\acrshort{mnl}}  & Train & 0.7250 & 0.7430 & 0.7247 & 0.7529 & 0.7917 \\
                                 & Test  & 0.7911 & 0.7924 & 0.7909 & 0.7924 & 0.8119 \\
\addlinespace
\multirow{2}{*}{\acrshort{dl}}   & Train & 0.6591 & 0.6740 & 0.6586 & 0.6840 & 0.7121 \\
                                 & Test  & 0.7227 & 0.7267 & 0.7221 & 0.7283 & 0.7448 \\
\addlinespace
\multirow{2}{*}{\acrshort{nl}}   & Train & 0.7235 & 0.7435 & 0.7214 & 0.7567 & 0.7910 \\
                                 & Test  & 0.7950 & 0.7961 & 0.7946 & 0.7962 & 0.8120 \\
\addlinespace
\multirow{2}{*}{\acrshort{cnl}}  & Train & 0.7233 & 0.7437 & 0.7239 & 0.7558 & 0.7895 \\
                                 & Test  & 0.7925 & 0.7936 & 0.7923 & 0.7938 & 0.8128 \\
\addlinespace
\multirow{2}{*}{\acrshort{lc}}   & Train & 0.7093 & 0.7404 & 0.8132 & 0.7658 & 0.7721 \\
                                 & Test  & 0.7875 & 0.7942 & 0.8354 & 0.7956 & 0.8107 \\
\addlinespace
\multirow{2}{*}{\acrshort{mxl}}  & Train & 0.7157 & 0.7230 & 0.7392 & 0.7614 & 0.7760 \\
                                 & Test  & 0.8007 & 0.8132 & 0.8049 & 0.8110 & 0.8206 \\
\addlinespace
\midrule
\multicolumn{7}{l}{\textit{Binary classification on the Swissmetro dataset}} \\ \midrule
\multirow{2}{*}{\acrshort{bp}}   & Train & 0.3176 & 0.3312 & 0.3164 & 0.3378 & 0.3914 \\
                                 & Test  & 0.3627 & 0.3652 & 0.3607 & 0.3787 & 0.4676 \\
\addlinespace
\midrule
\multicolumn{7}{l}{\textit{Ordinal classification on the Swissmetro dataset}} \\ \midrule
\multirow{2}{*}{\acrshort{ol}}   & Train & 0.8236 & 0.8334 & 0.8239 & 0.8369 & 0.8662 \\
                                 & Test  & 0.8536 & 0.8477 & 0.8540 & 0.8432 & 0.8627 \\
\addlinespace
 \midrule
\multicolumn{7}{l}{\textit{Multi-class classification on the Optima dataset}} \\ \midrule
\multirow{2}{*}{\acrshort{iclv}} & Train & 0.4582 & 0.4852 & 0.4526 & 0.5383 & 0.6028 \\
                                 & Test  & 0.5005 & 0.4822 & 0.4902 & 0.5047 & 0.5588 \\
\bottomrule
\end{tabular}
\end{table}

\FloatBarrier
\section{Additional figures}\label{app:figures}

\begin{figure}[htbp]
    \centering
    \begin{subfigure}[b]{0.48\textwidth}
        \centering
        \includegraphics[width=\textwidth]{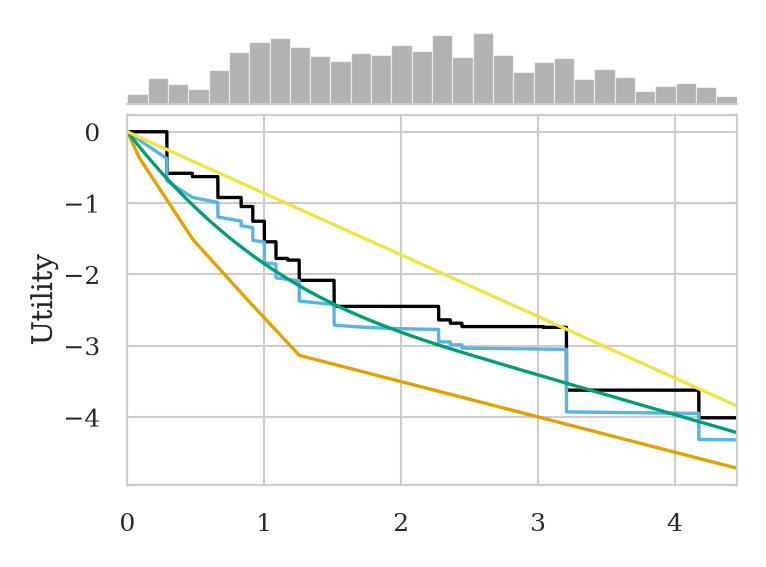}
        \caption{Train: travel time (hours)}
        \label{fig:dl_train_tt}
    \end{subfigure}
    \hfill
    \begin{subfigure}[b]{0.48\textwidth}
        \centering
        \includegraphics[width=\textwidth]{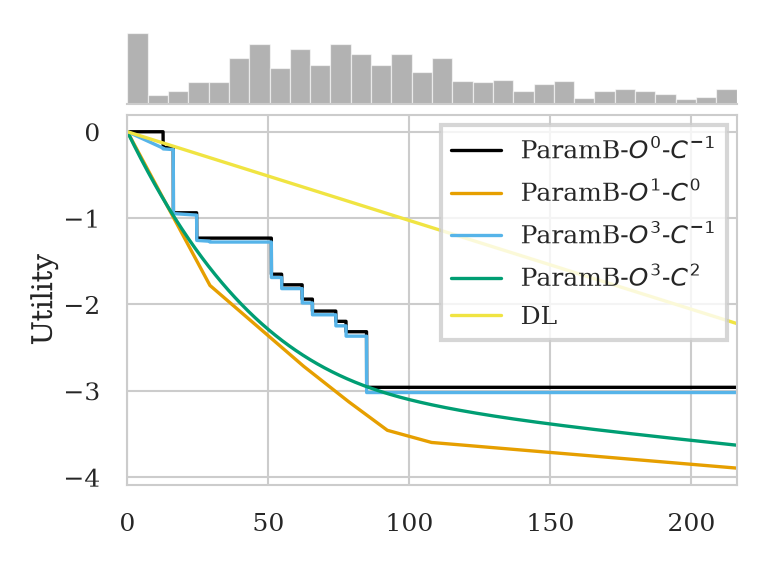}
        \caption{Train: cost (CHF)}
        \label{fig:dl_train_cost}
    \end{subfigure}

    \vspace{1em}
    \begin{subfigure}[b]{0.48\textwidth}
        \centering
        \includegraphics[width=\textwidth]{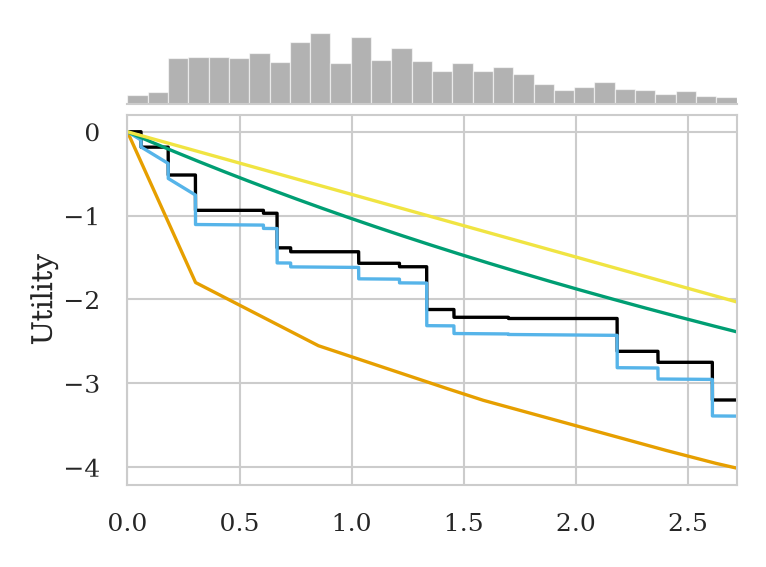}
        \caption{Swissmetro: travel time (hours)}
        \label{fig:dl_sm_tt}
    \end{subfigure}
    \hfill
    \begin{subfigure}[b]{0.48\textwidth}
        \centering
        \includegraphics[width=\textwidth]{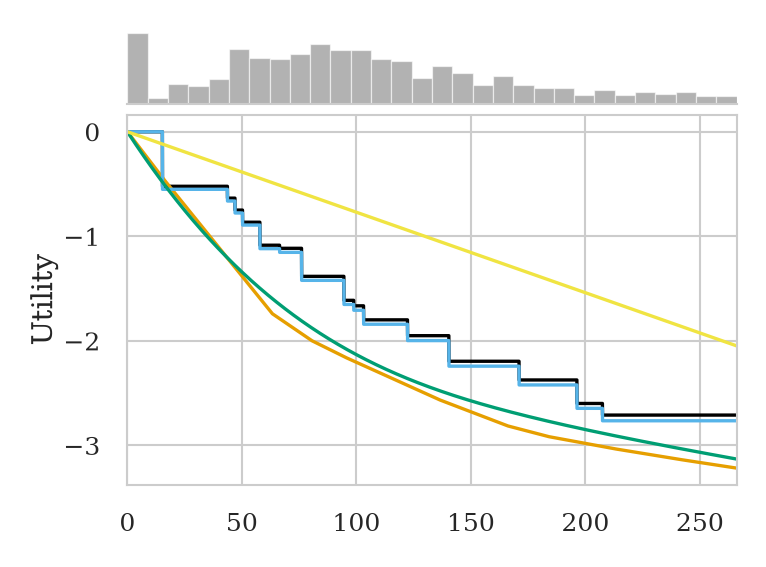}
        \caption{Swissmetro: cost (CHF)}
        \label{fig:dl_sm_cost}
    \end{subfigure}

    \vspace{1em}
    \begin{subfigure}[b]{0.48\textwidth}
        \centering
        \includegraphics[width=\textwidth]{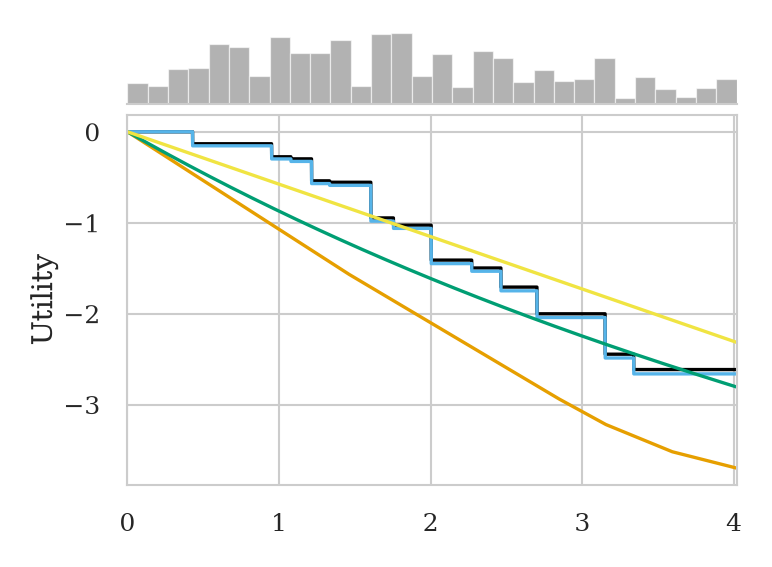}
        \caption{Car: travel time (hours)}
        \label{fig:dl_car_tt}
    \end{subfigure}
    \hfill
    \begin{subfigure}[b]{0.48\textwidth}
        \centering
        \includegraphics[width=\textwidth]{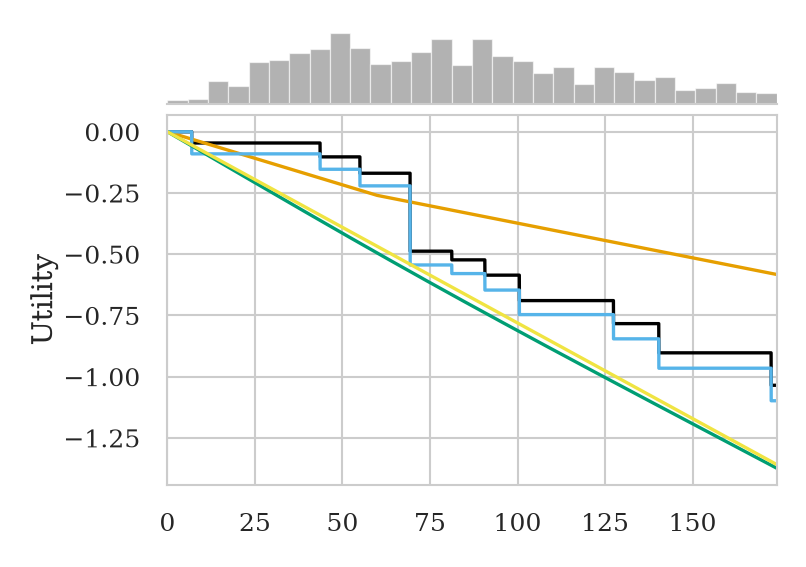}
        \caption{Car: cost (CHF)}
        \label{fig:dl_car_cost}
    \end{subfigure}

    \caption{Utility contributions of the trip attributes in the \acrshort{dl}, by alternative and model. Travel time is in hours and cost in CHF. Figures are cropped at the \nth{95} percentile of the feature, while its distribution is shown above the x-axis.}
    \label{fig:dl_utilities}
\end{figure}

\begin{figure}[htbp]
    \centering
    \begin{subfigure}[b]{0.48\textwidth}
        \centering
        \includegraphics[width=\textwidth]{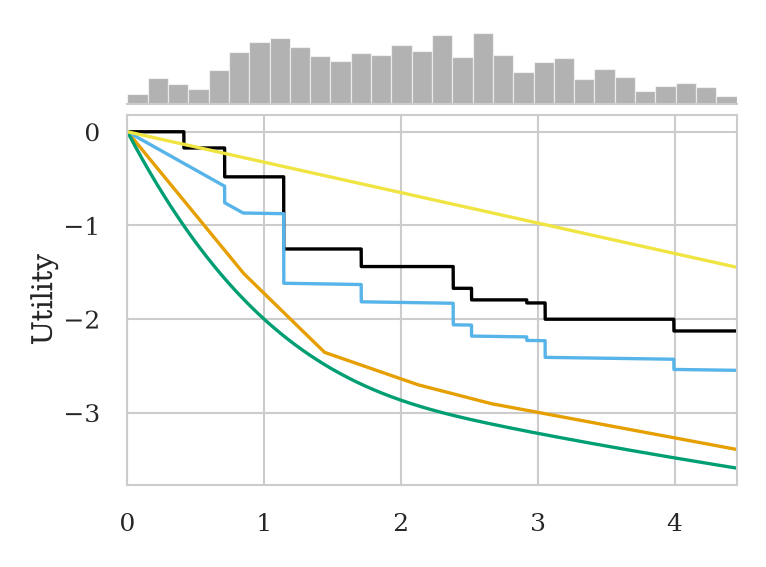}
        \caption{Train: travel time (hours)}
        \label{fig:bp_train_tt}
    \end{subfigure}
    \hfill
    \begin{subfigure}[b]{0.48\textwidth}
        \centering
        \includegraphics[width=\textwidth]{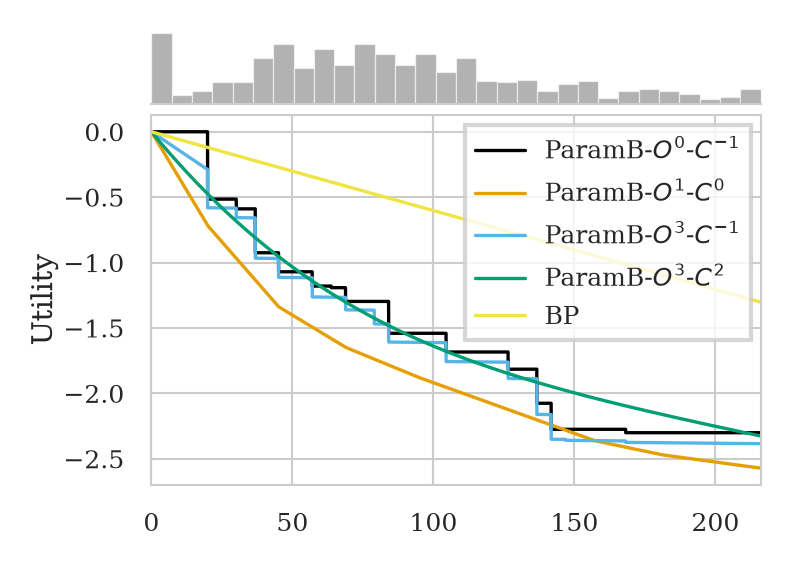}
        \caption{Train: cost (CHF)}
        \label{fig:bp_train_cost}
    \end{subfigure}

    \vspace{1em}
    \begin{subfigure}[b]{0.48\textwidth}
        \centering
        \includegraphics[width=\textwidth]{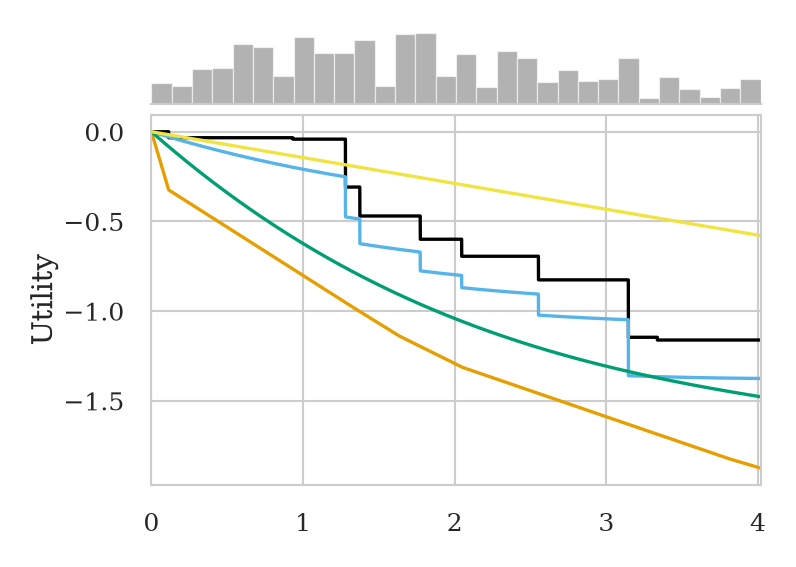}
        \caption{Car: travel time (hours)}
        \label{fig:bp_car_tt}
    \end{subfigure}
    \hfill
    \begin{subfigure}[b]{0.48\textwidth}
        \centering
        \includegraphics[width=\textwidth]{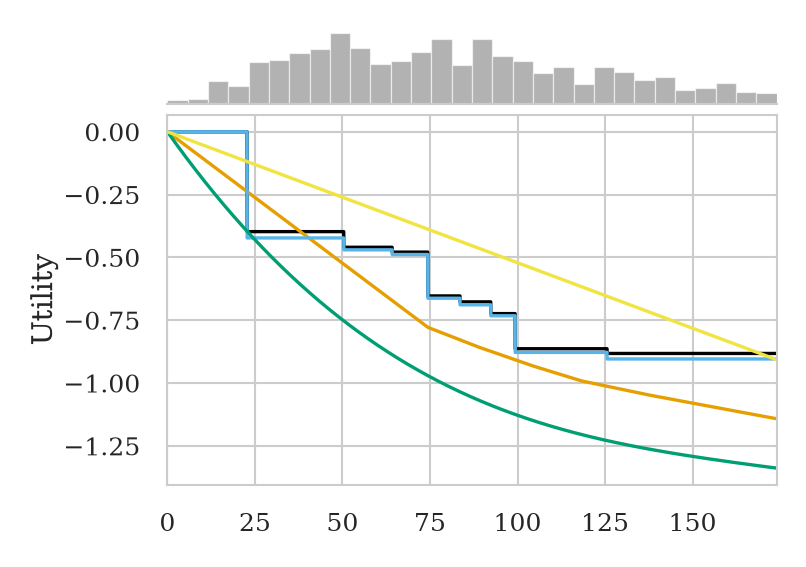}
        \caption{Car: cost (CHF)}
        \label{fig:bp_car_cost}
    \end{subfigure}
    \caption{Utility contributions of the trip attributes in the \acrshort{bp}, by model. Travel time is in hours and cost in CHF. Figures are cropped at the \nth{95} percentile of the feature, while its distribution is shown above the x-axis.}
    \label{fig:bp_utilities}
\end{figure}

\begin{figure}[htbp]
    \centering
    \begin{subfigure}[b]{0.48\textwidth}
        \centering
        \includegraphics[width=\textwidth]{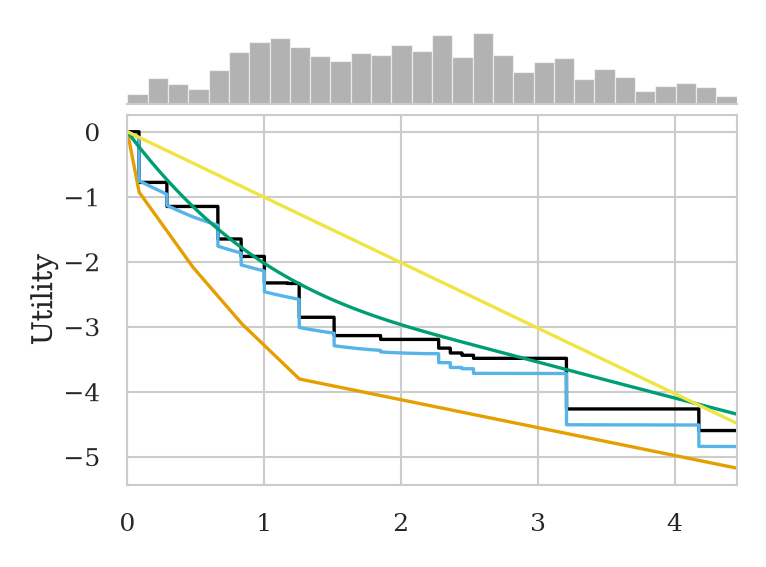}
        \caption{Train: travel time (hours)}
        \label{fig:mnl_train_tt}
    \end{subfigure}
    \hfill
    \begin{subfigure}[b]{0.48\textwidth}
        \centering
        \includegraphics[width=\textwidth]{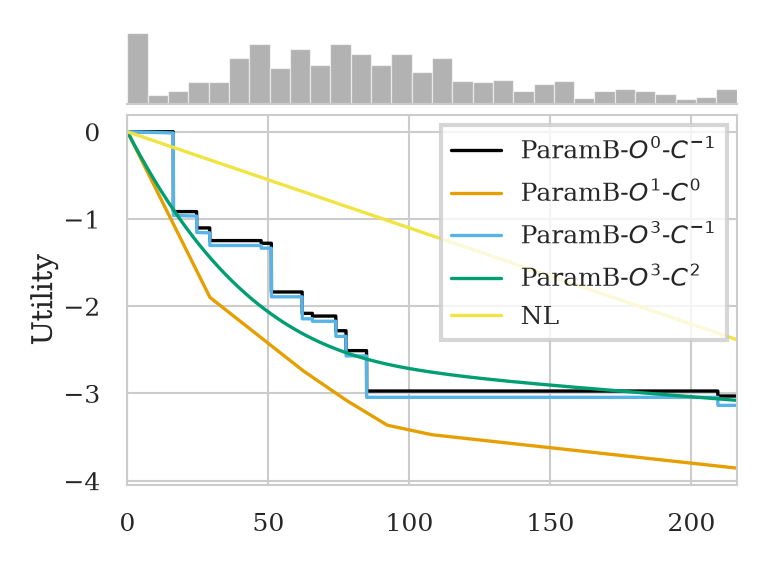}
        \caption{Train: cost (CHF)}
        \label{fig:mnl_train_cost}
    \end{subfigure}

    \vspace{1em}
    \begin{subfigure}[b]{0.48\textwidth}
        \centering
        \includegraphics[width=\textwidth]{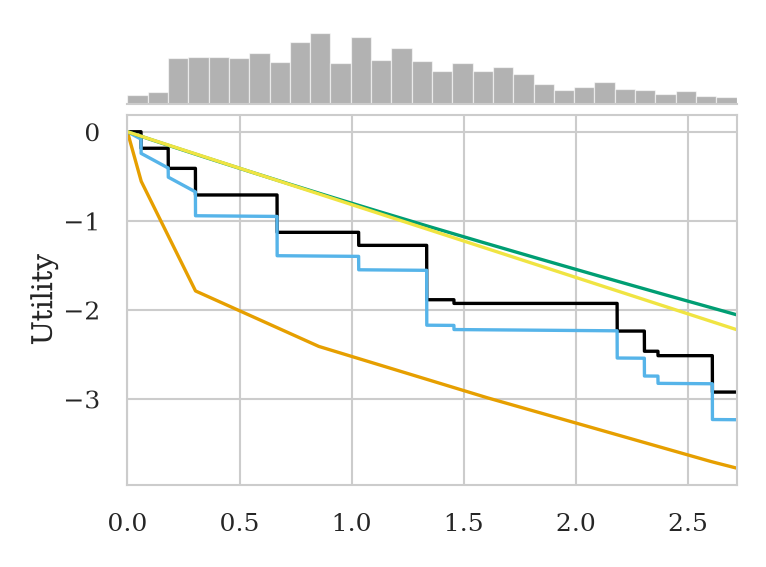}
        \caption{Swissmetro: travel time (hours)}
        \label{fig:mnl_sm_tt}
    \end{subfigure}
    \hfill
    \begin{subfigure}[b]{0.48\textwidth}
        \centering
        \includegraphics[width=\textwidth]{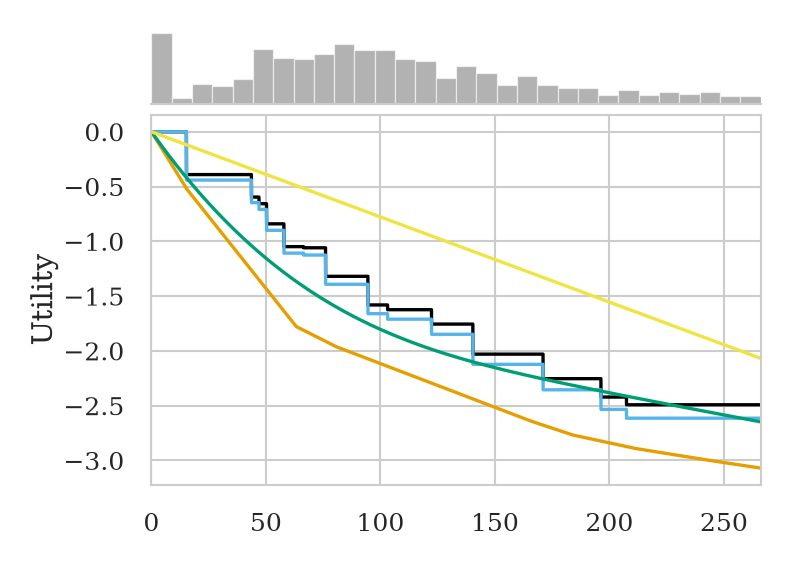}
        \caption{Swissmetro: cost (CHF)}
        \label{fig:mnl_sm_cost}
    \end{subfigure}

    \vspace{1em}
    \begin{subfigure}[b]{0.48\textwidth}
        \centering
        \includegraphics[width=\textwidth]{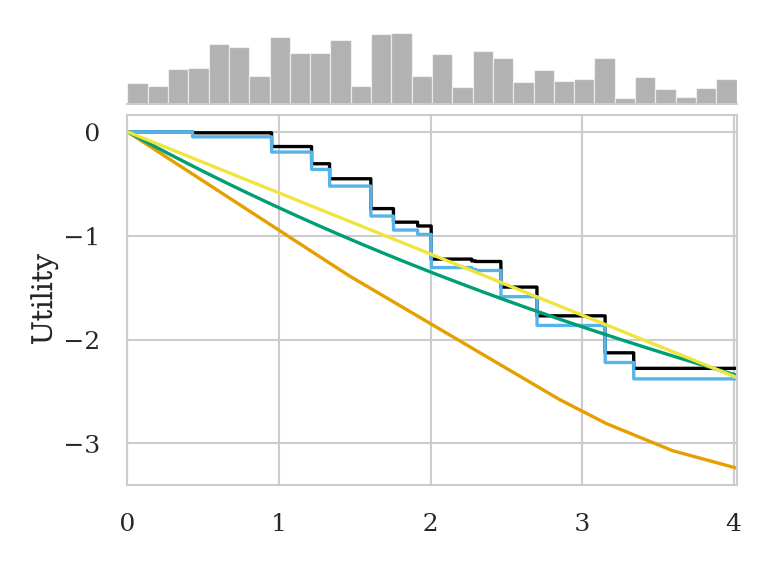}
        \caption{Car: travel time (hours)}
        \label{fig:mnl_car_tt}
    \end{subfigure}
    \hfill
    \begin{subfigure}[b]{0.48\textwidth}
        \centering
        \includegraphics[width=\textwidth]{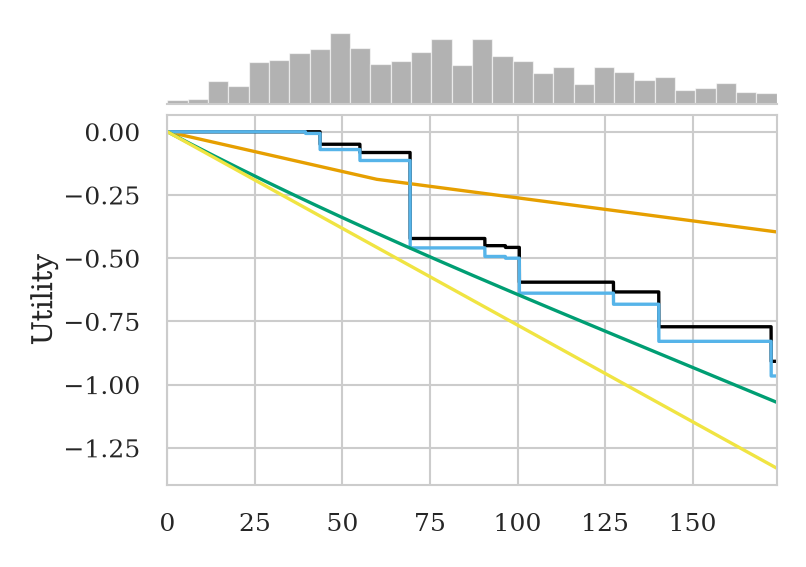}
        \caption{Car: cost (CHF)}
        \label{fig:mnl_car_cost}
    \end{subfigure}

    \caption{Utility contributions of the trip attributes in the \acrshort{nl}, by alternative and model. Travel time is in hours and cost in CHF. Figures are cropped at the \nth{95} percentile of the feature, while its distribution is shown above the x-axis.}
    \label{fig:nl_utilities}
\end{figure}

\begin{figure}[htbp]
    \centering
    \begin{subfigure}[b]{0.48\textwidth}
        \centering
        \includegraphics[width=\textwidth]{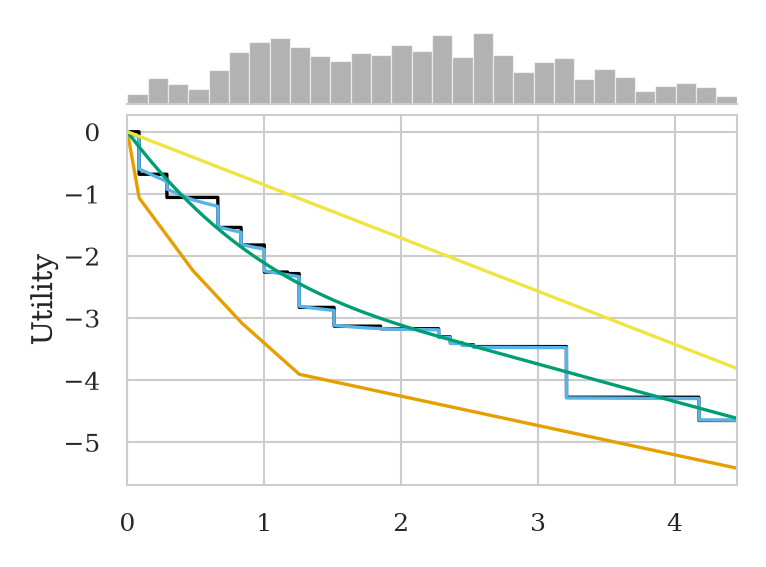}
        \caption{Train: travel time (hours)}
        \label{fig:mcnl_train_tt}
    \end{subfigure}
    \hfill
    \begin{subfigure}[b]{0.48\textwidth}
        \centering
        \includegraphics[width=\textwidth]{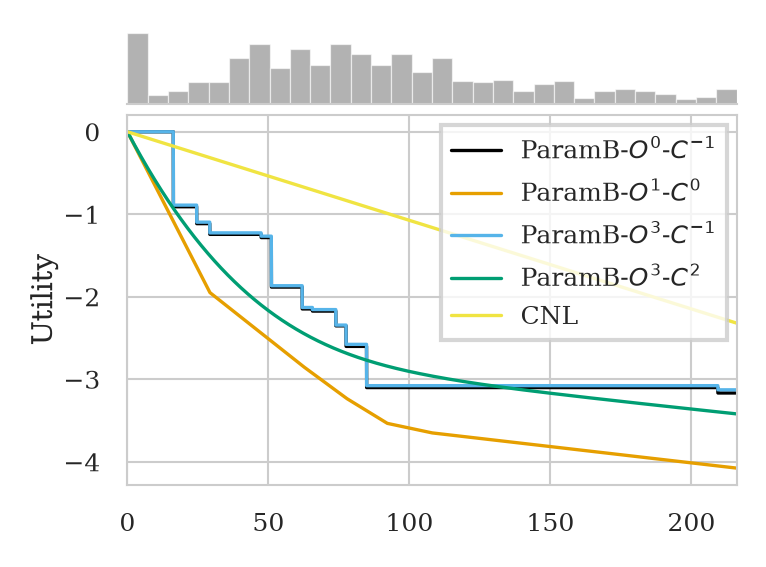}
        \caption{Train: cost (CHF)}
        \label{fig:mcnl_train_cost}
    \end{subfigure}

    \vspace{1em}
    \begin{subfigure}[b]{0.48\textwidth}
        \centering
        \includegraphics[width=\textwidth]{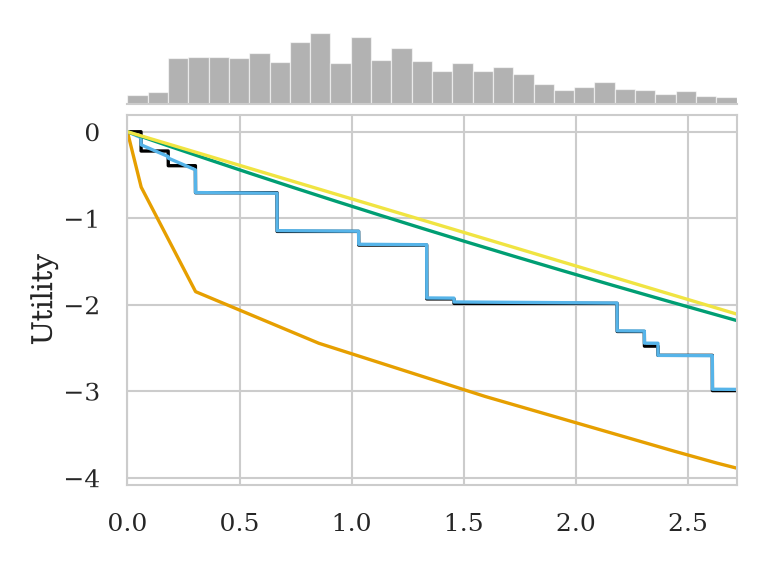}
        \caption{Swissmetro: travel time (hours)}
        \label{fig:mcnl_sm_tt}
    \end{subfigure}
    \hfill
    \begin{subfigure}[b]{0.48\textwidth}
        \centering
        \includegraphics[width=\textwidth]{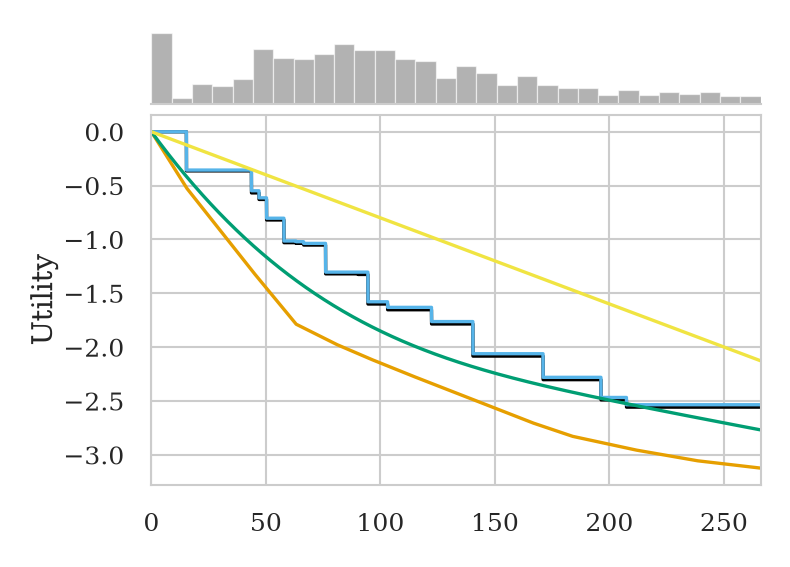}
        \caption{Swissmetro: cost (CHF)}
        \label{fig:mcnl_sm_cost}
    \end{subfigure}

    \vspace{1em}
    \begin{subfigure}[b]{0.48\textwidth}
        \centering
        \includegraphics[width=\textwidth]{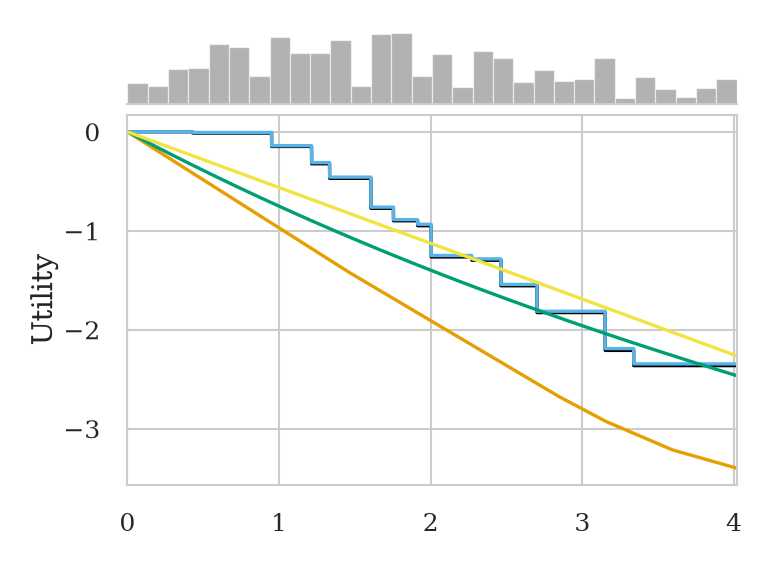}
        \caption{Car: travel time (hours)}
        \label{fig:mcnl_car_tt}
    \end{subfigure}
    \hfill
    \begin{subfigure}[b]{0.48\textwidth}
        \centering
        \includegraphics[width=\textwidth]{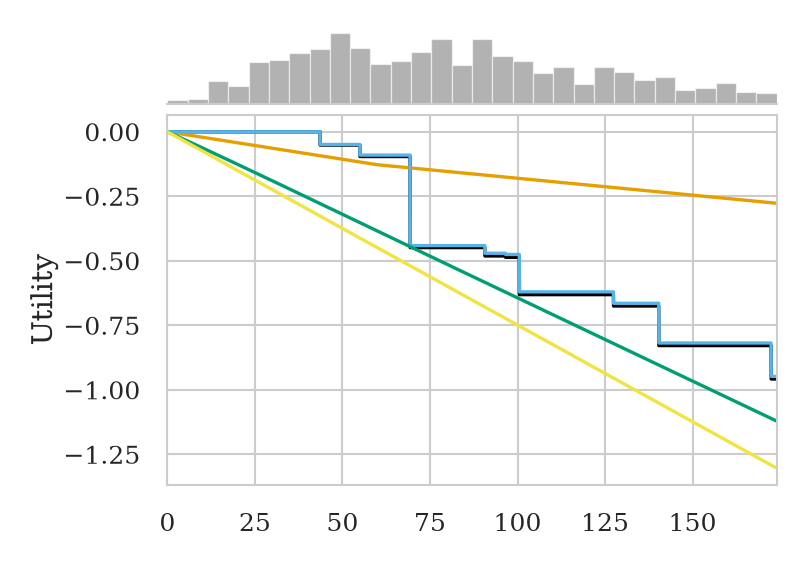}
        \caption{Car: cost (CHF)}
        \label{fig:mcnl_car_cost}
    \end{subfigure}

    \caption{Utility contributions of the trip attributes in the \acrshort{cnl}, by alternative and model. Travel time is in hours and cost in CHF. Figures are cropped at the \nth{95} percentile of the feature, while its distribution is shown above the x-axis.}
    \label{fig:cnl_utilities}
\end{figure}

\begin{figure}[htbp]
    \centering
    \begin{subfigure}[b]{0.32\textwidth}
        \centering
        \includegraphics[width=\textwidth]{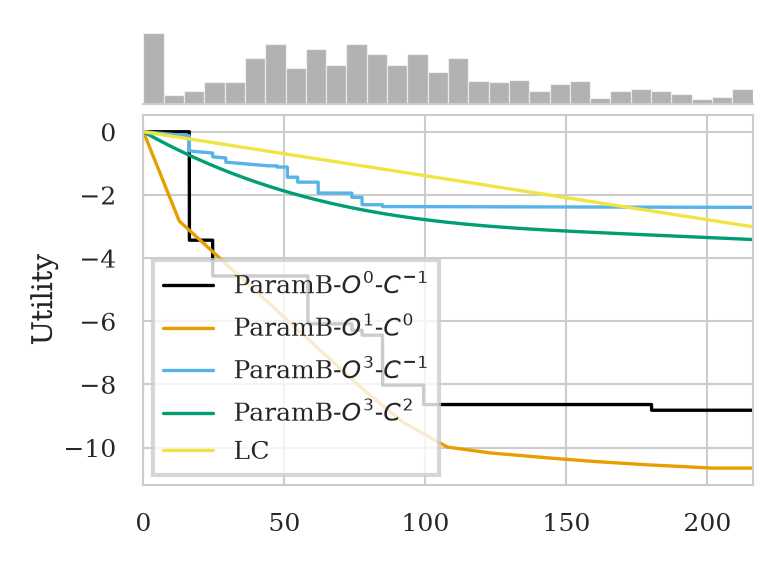}
        \caption{Train: cost (class 1) (CHF)}
        \label{fig:dl_train_tt}
    \end{subfigure}
    \hfill
    \begin{subfigure}[b]{0.32\textwidth}
        \centering
        \includegraphics[width=\textwidth]{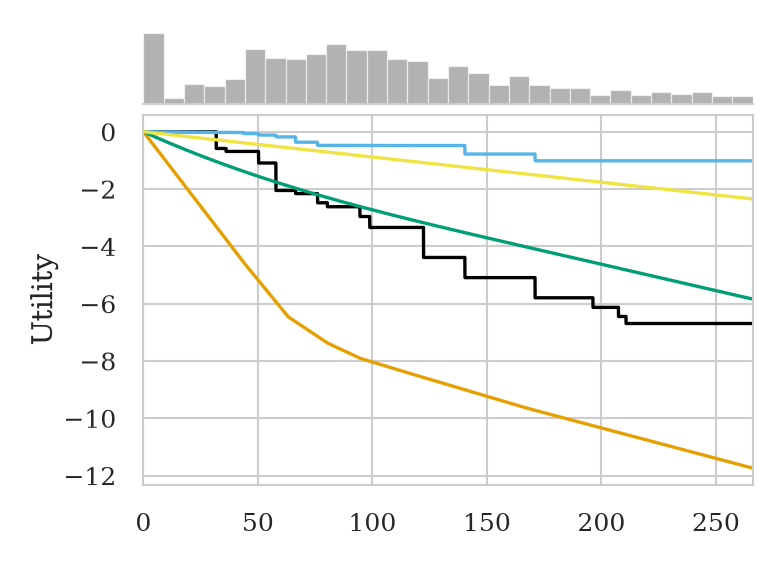}
        \caption{Swissmetro: cost (class 1) (CHF)}
        \label{fig:dl_train_tt}
    \end{subfigure}
    \hfill
    \begin{subfigure}[b]{0.32\textwidth}
        \centering
        \includegraphics[width=\textwidth]{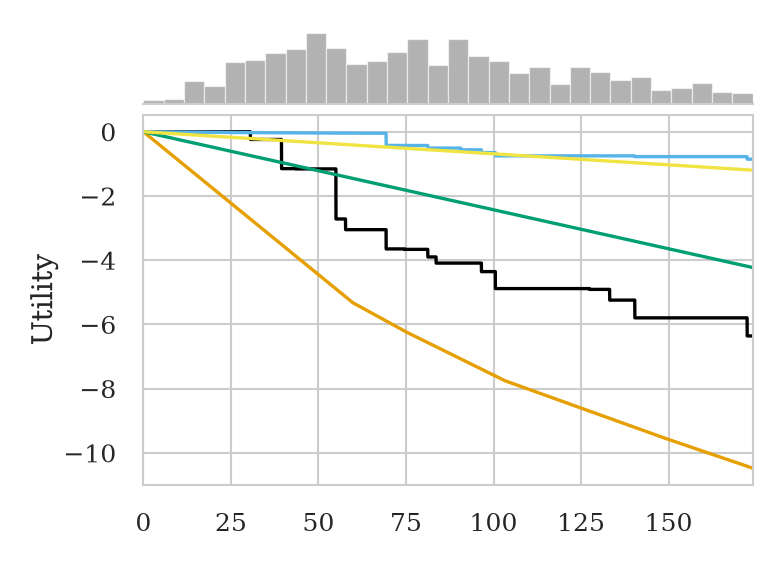}
        \caption{Car: cost (class 1) (CHF)}
        \label{fig:dl_train_tt}
    \end{subfigure}
    \vspace{1em}
    \begin{subfigure}[b]{0.32\textwidth}
        \centering
        \includegraphics[width=\textwidth]{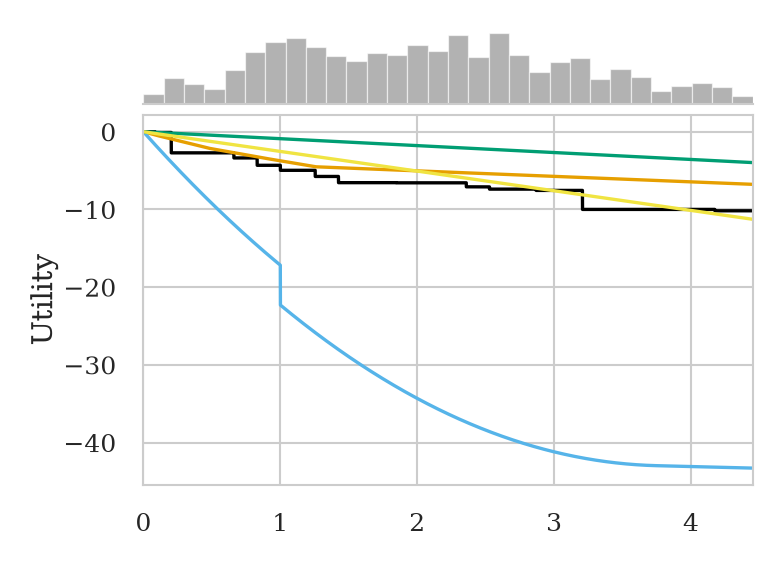}
        \caption{Train: travel time (class 2) (hours)}
        \label{fig:dl_train_cost}
    \end{subfigure}
    \hfill
    \begin{subfigure}[b]{0.32\textwidth}
        \centering
        \includegraphics[width=\textwidth]{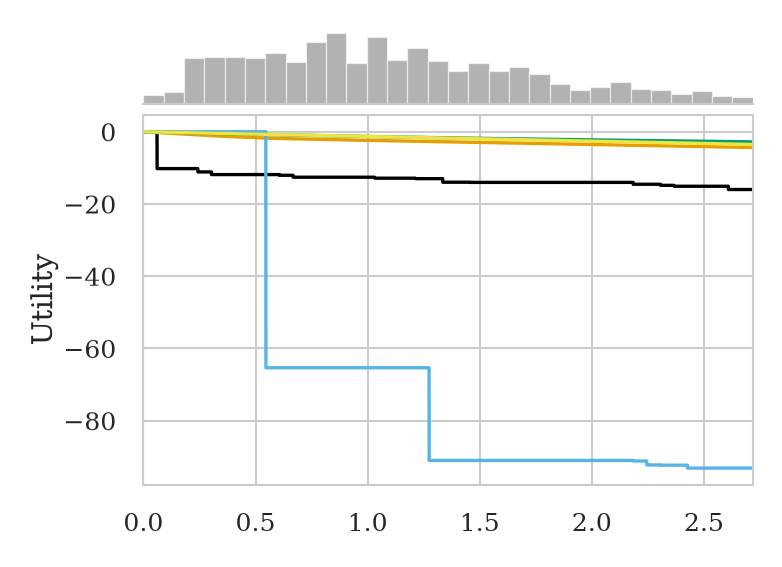}
        \caption{Swissmetro: travel time (class 2) (hours)}
        \label{fig:dl_sm_tt}
    \end{subfigure}
    \hfill
    \begin{subfigure}[b]{0.32\textwidth}
        \centering
        \includegraphics[width=\textwidth]{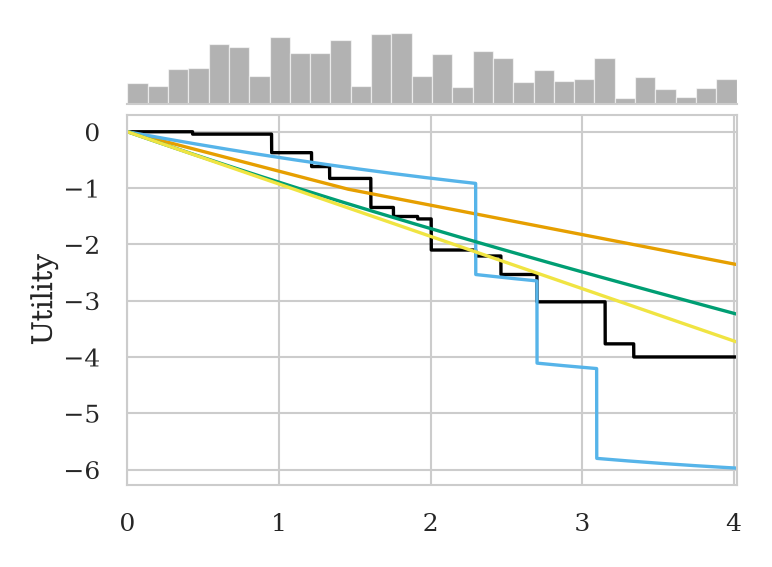}
        \caption{Car: travel time (class 2) (hours)}
        \label{fig:dl_car_tt}
    \end{subfigure}

    \vspace{1em}
    \begin{subfigure}[b]{0.32\textwidth}
        \centering
        \includegraphics[width=\textwidth]{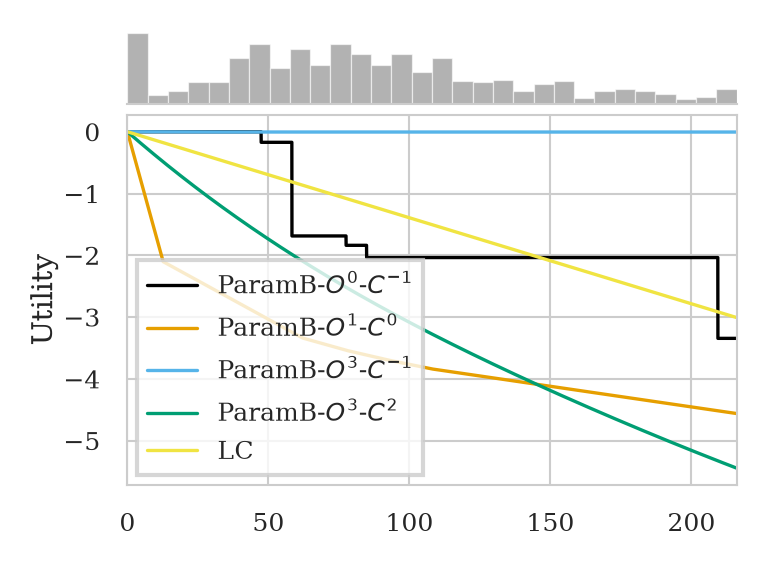}
        \caption{Train: cost (class 2) (CHF)}
        \label{fig:dl_train_cost}
    \end{subfigure}
    \hfill
    \begin{subfigure}[b]{0.32\textwidth}
        \centering
        \includegraphics[width=\textwidth]{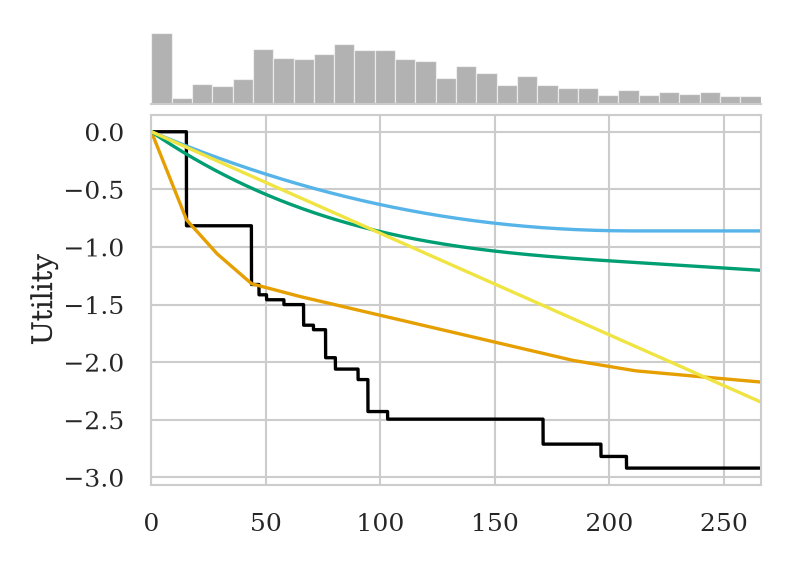}
        \caption{Swissmetro: cost (class 2) (CHF)}
        \label{fig:dl_sm_cost}
    \end{subfigure}
    \hfill
    \begin{subfigure}[b]{0.32\textwidth}
        \centering
        \includegraphics[width=\textwidth]{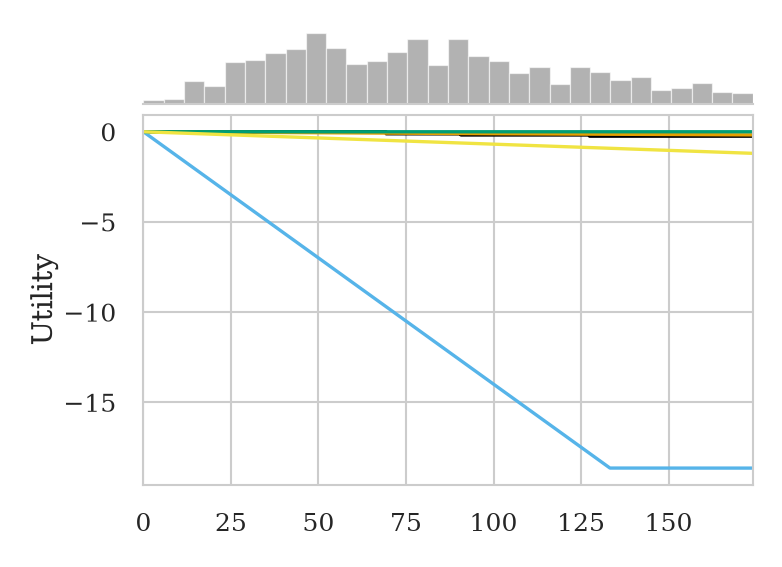}
        \caption{Car: cost (class 2) (CHF)}
        \label{fig:dl_car_cost}
    \end{subfigure}

    \caption{Utility contributions of the trip attributes in the \acrshort{lc}, by alternative and model. Travel time is in hours and cost in CHF. Figures are cropped at the \nth{95} percentile of the feature, while its distribution is shown above the x-axis.}
    \label{fig:lc_utilities}
\end{figure}

\FloatBarrier
\section{Structural parameters}\label{app:structural}

\Cref{tab:structural} compares the structural parameters of the \acrshort{nl}, \acrshort{cnl} and \acrshort{lc} models. For the \acrshort{nl}, all models estimate similar nest parameters for the existing alternatives, between 1.23 and 1.38, which imply a correlation of 0.34 to 0.47 between the utilities of train and car ($1 - 1/\mu^2$).

For the \acrshort{cnl}, the \acrshort{paramboost} variants estimate a public transport nest parameter between 1.17 and 1.26. Three of the four estimate a membership of train in the existing nest of about 0.42, matching the baseline, while \pb{3}{2} gives 0.28. The nest parameter of the existing alternatives is fixed to one in the \acrshort{paramboost} models but estimated at 1.43 in the baseline, which also estimates $\mu_\text{Public}$ at 3.00 (upper bound); this difference in normalisation limits direct comparison of the nest parameters.

For the \acrshort{lc} model, the baseline assigns a probability of 0.12 to class 1, and the \acrshort{paramboost} variants range from 0.29 to 0.36, except \pb{3}{-1} (0.76), the unstable variant identified in \cref{sec:performance}. This means the \acrshort{paramboost} variants' probabilities are more spread across the two classes than the linear-in-parameters baseline.

\begin{table}[htbp]
\centering
\caption{Estimated structural parameters. For the \acrshort{cnl}, $\alpha$ is the membership of train in the nest of existing alternatives; for the \acrshort{lc}, $\pi$ is the probability of belonging to class 1.}
\label{tab:structural}
\footnotesize
\begin{tabular}{lrrrrr}
\toprule
 & \acrshort{nl} & \multicolumn{3}{c}{\acrshort{cnl}} & \acrshort{lc} \\
\cmidrule(lr){2-2}\cmidrule(lr){3-5}\cmidrule(lr){6-6}
 & $\mu_\text{Existing}$ & $\mu_\text{Existing}$ & $\mu_\text{Public}$ & $\alpha$ & $\pi$ \\
\midrule
\pb{0}{-1} & 1.264 & 1.000 & 1.184 & 0.425 & 0.294 \\
\pb{1}{0}  & 1.230 & 1.000 & 1.174 & 0.416 & 0.361 \\
\pb{3}{-1} & 1.257 & 1.000 & 1.185 & 0.425 & 0.759 \\
\pb{3}{2}  & 1.380 & 1.000 & 1.257 & 0.276 & 0.287 \\
Linear     & 1.274 & 1.426 & 3.000 & 0.420 & 0.123 \\
\bottomrule
\end{tabular}
\end{table}

\FloatBarrier
\bibliographystyle{elsarticle-harv}
\bibliography{cas-refs}

\end{document}